\documentclass[letterpaper]{article} %
\usepackage{aaai2027}
\nocopyright  %
\usepackage[hyphens]{url}  %
\usepackage{graphicx} %
\usepackage{natbib}  %
\usepackage{caption} %
\usepackage{algorithm}
\usepackage{algorithmic}
\usepackage{booktabs}
\usepackage{afterpage}
\usepackage{amsmath}
\usepackage{amssymb}
\usepackage{array}
\usepackage{multirow}
\usepackage{nicematrix}
\usepackage{adjustbox}
\usepackage{fancyvrb}
\usepackage{comment}
\usepackage{enumitem}
\usepackage{xcolor}

\newcommand{\first}[1]{\textbf{#1}}

\graphicspath{{figs/}}

\title{ReasonCast: Towards Explainable Time Series Forecasting with Reasoning}

\author{
    Seunghan Lee,
    Jun Seo,
    Jaehoon Lee,
    Junhyeok Kang,
    Sangjun Han,
    Sungdong Yoo,\\
    Minjae Kim,
    Tae Yoon Lim,
    Dongwan Kang,
    Hwanil Choi,
    Soonyoung Lee,
    Wonbin Ahn
}
\affiliations{
    LG AI Research
}

\begin{document}

\maketitle

\begin{abstract}
Most time series (TS) models are specialized for a single task, 
either \textit{understanding} (i.e., returning text answers about a TS) 
or \textit{generation} (i.e., returning a numeric forecast). 
Only recently have unified models begun to handle the two within a single architecture. 
Even these models, however, produce the two outputs as task-separated paths and 
\textit{cannot \textbf{predict} a series and \textbf{explain} why that prediction arises} within a \textit{single} coherent response. 
In this paper, we argue for a task-fused model that \textit{jointly} produces 1) prediction (generation) and 2) self-explanation (understanding),
thereby integrating 1) numerical TS forecasting and 2) interpretable text reasoning within a \textit{single} response.
To enable the systematic study of this capability, we present both a benchmark and a recipe
that jointly address 
the two tasks.
The benchmark, \textbf{ReasonTS-Bench}, identifies five fundamental patterns underlying TS 
and enables the joint evaluation of 
both tasks.
\textbf{ReasonCast}, our recipe for fine-tuning any LLM 
to perform both tasks jointly,
yields a model that generates a 
reasoning chain and a
forecast together in a single autoregressive pass.
Extensive experiments show that ReasonCast outperforms 
both LLMs and 
TS models
on prediction accuracy while 
producing verifiable, causal reasoning.
Code is available at: \url{https://github.com/seunghan96/reasoncast}.
\end{abstract}

\section{Introduction}

Time series (TS) models have traditionally been built for forecasting by generating future numeric values from past observations (i.e., generation (\textbf{G}))~\citep{Nie2023PatchTST,TSFM}.
With the rise of large language models (LLMs), a line of work emerged that explains or reasons about a series in natural language, focusing on text understanding rather than TS generation (i.e., understanding (\textbf{U}))~\citep{TSLM,TSRM}. 
More recently, unified models have begun to perform both within a single architecture (i.e., understanding + generation (\textbf{U+G}))~\citep{TimeOmniVL}.

Even so, \textit{no existing model performs the two jointly},
that is, forecasting a series and explaining that forecast
\textit{in a single response} (i.e., Understanding × Generation (\textbf{UxG})).
We argue that this task is a key requirement for explainable TS AI, as many real-world applications require users to understand the rationale behind a forecast before acting upon it.
This paper aims to fill that gap by introducing a \textit{unified framework} that forecasts future values while simultaneously explaining the rationale behind its predictions.

\begin{table}[t]
\centering
\adjustbox{max width=\columnwidth}{%
\begin{tabular}{@{}cl|>{\centering\arraybackslash}p{0.34\columnwidth}|>{\centering\arraybackslash}p{0.34\columnwidth}@{}}
\toprule
 & \textbf{Axis} & Understanding & Generation \\
\midrule
\midrule
(1) & U \;\, & $p_\theta(o_{\mathrm{und}} \mid X)$ & --- \\
\midrule
(2) & G \;\, & --- & $p_\theta(o_{\mathrm{gen}} \mid X)$ \\
\midrule
(3) & U {+} G \;\, & $p_\theta(o_{\mathrm{und}} \mid X)$ &  $p_\theta(o_{\mathrm{gen}} \mid X)$ \\
\midrule
(4) & \textbf{U} $\times$ \textbf{G} \;\, & \multicolumn{2}{c}{$\boldsymbol{p_\theta(o_{\mathrm{gen}}, o_{\mathrm{rsn}} \mid X)}$} \\
\bottomrule
\end{tabular}}
\caption{
U $\times$ G framework. 
We propose a unified framework that generates forecasts and 
explanations within a single pass.} 

\label{tab:axes}
\end{table}

To this end, we formulate forecasting and explanation as a \textit{single} autoregressive process, as shown in Table~\ref{tab:axes}. 
Formally, the model first 1) emits a reasoning chain $o_{\mathrm{rsn}}$ and then 2) generates the forecast $o_{\mathrm{gen}}$ conditioned on it as:
\begin{equation}
p_\theta(o_{\mathrm{gen}}, o_{\mathrm{rsn}} \mid X)
= p_\theta(o_{\mathrm{rsn}} \mid X) \cdot
  p_\theta(o_{\mathrm{gen}} \mid o_{\mathrm{rsn}}, X),
\label{eq:fwdfact}
\end{equation}
so the explanation \textit{causally} guides the prediction rather than serving as a post-hoc justification.
In this paper, we introduce both a benchmark and a training recipe under this framework.
The benchmark, \textbf{ReasonTS-Bench}, is a controlled synthetic dataset built on five primitives that isolate elementary components of TS
(e.g., periodicity, trend, temporal dependence).\footnote{These five primitives are sufficient and widely used, forming the basis of TS decomposition~\citep{Cleveland1990STL}.}
\textbf{ReasonCast}, our recipe for \textit{any} LLM, is the first to generate a forecast and its explanation within a single autoregressive pass.
Our main contributions are:
\setlist[itemize]{leftmargin=8pt,itemsep=1pt,topsep=1pt}
\begin{itemize}
\item \textbf{Framework.} 
We propose a unified framework that formulates forecasting and explanation as a single coherent task.
\item \textbf{Benchmark.} 
We introduce ReasonTS-Bench, a controlled benchmark with five disentangled TS patterns
and ground-truth reasoning, so that prediction and explanation can be scored jointly
rather than by forecast error alone.
\item \textbf{Recipe.}
We propose ReasonCast, a training recipe that yields the first model to jointly generate a forecast and its underlying explanation within a single autoregressive pass.
\item \textbf{Experiments.} 
ReasonCast outperforms TS and LLM-based baselines on TS forecasting with reasoning.
\end{itemize}

\begin{table*}[t]
\centering
\setlength{\tabcolsep}{2pt}
\renewcommand{\arraystretch}{1.25}

\small
\begin{tabular}{@{}cl|>{\raggedright\arraybackslash}p{0.55\textwidth}|>{\raggedright\arraybackslash}p{0.33\textwidth}@{}}
\toprule
 & \textbf{Axis} & \textbf{[1] Representative models} & \textbf{[2] TS benchmarks} \\
\midrule
(1) & U \;\,
    & ChatTS, Time-MQA, S2TS-LLM, Thoth, PATRA
    & TSQA, MTBench, SenTSR-Bench \\
(2) & G \;\,
    & DLinear, PatchTST, Chronos, Moirai, TimesFM, VisionTS, Time-MoE
    & Monash, GIFT-Eval, TFB \\
(3) & U {+} G \;\,
    & TimeOmni-1, ChatTime, TimeOmni-VL
    & TSR-Suite, FinSTaR \\
(4) & \textbf{U} $\times$ \textbf{G} \;\,
    & LLMs $+$ \textbf{ReasonCast (Ours)}
    & \textbf{ReasonTS-Bench (Ours)} \\
\bottomrule
\end{tabular}
\caption{Related works by axis. Representative TS models and benchmarks grouped by the four axes of Table~\ref{tab:axes}. 
Only (4) forecasts and explains why within a single pass, and scores the forecast and its reasoning chain jointly.
}
\label{tab:related_all}
\end{table*}

\begin{figure*}[t]
\centering
\adjustbox{max width=\textwidth}{\includegraphics{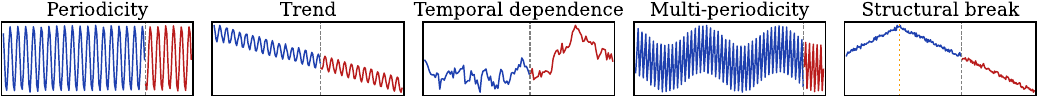}}
\caption{Five primitives of ReasonTS-Bench. One sample per pattern. Blue is the
input and red is the target, and the ground-truth reasoning excerpt paired
with each sample is given in Appendix~\ref{sec:appendix}.
Each pattern uses a different reasoning schema.}
\label{fig:bench}
\end{figure*}

\section{Related Work}
\textbf{TS forecasting models (Generation).} 
These models predict numeric futures without a textual interface and are typically trained on individual datasets.
DLinear~\citep{Zeng2023DLinear} 
performs TS forecasting with a linear model,
PatchTST~\citep{Nie2023PatchTST} patches the series into 
tokens, TimesNet~\citep{TimesNet} models 2D period variation, and
iTransformer~\citep{Liu2024iTransformer} attends across variates. Foundation
models instead pretrain across datasets: Chronos~\citep{TSFM} tokenizes values
into a language, Moirai~\citep{Moirai} trains one universal forecaster across
domains, TimesFM~\citep{TimesFM} is a decoder-only forecaster, VisionTS
\citep{VisionTS} casts forecasting as image inpainting, and Time-MoE
\citep{TimeMoE} scales with mixture-of-experts. Time-LLM~\citep{TimeLLM} instead
reprograms a frozen LLM to forecast. 
In all cases, the output is numeric future values \textit{with no textual explanation}.
\textbf{TS reasoning models (Understanding).} 
These models take a series and a textual query and return a textual answer about the series, not a numeric forecast.
ChatTS~\citep{TSLM} aligns a series with an LLM through synthetic data to answer questions about it.
Other work probes whether LLMs can describe and explain temporal dynamics in language~\citep{TSRM}.
Time-MQA~\citep{TimeMQA} continually pretrains an LLM for multi-task question answering over series, 
and S2TS-LLM~\citep{S2TSLLM} symbolizes a series so an LLM can analyze it. 
More recent work mid-trains an LLM to bridge it to TS understanding (Thoth~\citep{Thoth}) 
and aligns temporal patterns for question answering (PATRA~\citep{PATRA}). 
Their outputs are text, not numbers, \textit{so they cannot forecast the future signal}.

\textbf{Unified TS models (Understanding + Generation).} 
TimeOmni-1~\citep{TimeOmni1}
incentivizes TS reasoning in an LLM, covering perception and event-aware
forecasting as separate 
tasks. ChatTime~\citep{ChatTime} is a 
TS
foundation model that takes and emits both numbers and text, 
but through separate queries. FinSTaR~\citep{FinSTaR} 
is a financial TS reasoning model that couples a reasoning trace with a prediction but treats the two as separate tasks. 
TimeOmni-VL~\citep{TimeOmniVL}
renders the series as an image and trains a vision-language backbone
to handle both 
tasks.
Its two paths share parameters but are conditionally independent given the
input, so a user must pick one prompt at a time. These models form the U+G category in
Table~\ref{tab:axes}, the closest prior work to ours, which unify understanding
and generation \textit{but still solve each as its own task}, so the reasoning serves
task performance rather than grounding a forecast within a single coherent
output. A concurrent agentic line, such as AlphaCast~\citep{AlphaCast}, also
pairs reasoning with forecasting, but orchestrates a frozen LLM through a
multi-turn workflow rather than a single model.

\textbf{TS benchmarks and symbolic regression.} Two further threads are covered in
Appendix~\ref{sec:appendix_relatedwork}: how existing TS benchmarks map onto the four
axes, and how symbolic regression relates to a model that both predicts and explains.
Table~\ref{tab:related_all} gives the full comparison of representative models and benchmarks across all four axes.

\section{ReasonTS-Bench}
\label{sec:bench}

\begin{figure*}[t]
\centering
\adjustbox{max width=\textwidth}{\includegraphics{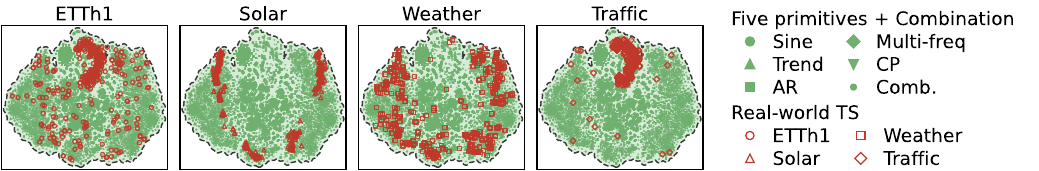}}
\caption{Sufficiency of the five primitives. Real-world datasets lie within the region occupied by the five primitives.}
\label{fig:coverage}
\end{figure*}

Existing TS benchmarks are not designed to evaluate whether forecasts and explanations are mutually consistent. 
Prediction and explanation metrics are typically measured in isolation, and real-world TS datasets provide neither ground-truth reasoning nor a way to verify it. 
To address these limitations, we propose \textbf{ReasonTS-Bench}, a synthetic benchmark with algorithmically generated and verifiable reasoning targets.

\subsection{Primitives of TS}

The benchmark is built from five primitives, shown in Figure~\ref{fig:bench}, each isolating one elementary building block of TS:

\setlist[itemize]{leftmargin=8pt,itemsep=1pt,topsep=1pt}
\begin{itemize}
\item \textbf{Periodicity}: A single sinusoid.
\item \textbf{Trend}: A sinusoid plus a deterministic linear trend.
\item \textbf{Temporal dependence}: An AR(1) process.
\item \textbf{Multi-periodicity}: A superposition of two frequencies.
\item \textbf{Structural break}: A changepoint in the dynamics.
\end{itemize}
\textbf{Sufficiency of the five primitives.}
Although ReasonTS-Bench is synthetic, its five primitives are not arbitrary.
Together, they form \textit{the canonical structures of TS}~\citep{Wold1938,Cleveland1990STL}, and are used to pretrain TS foundation models~\citep{TSFM}. Because every series is generated from known parameters, we can vary one structure 
while holding the rest fixed, a control that natural corpora do not offer.
To test this beyond synthetic data, we embed windows from four real-world datasets alongside the primitives with t-SNE, and Figure~\ref{fig:coverage} shows they lie within the primitive region, 
suggesting the primitives cover real-world structures.
Table~\ref{tab:bench} lists the equation used to generate each pattern, Figure~\ref{fig:bench} shows one example of each pattern, and Figure~\ref{fig:worked} gives a full ground-truth reasoning chain for one of them.
In the results, we refer to each primitive by the name of its generating equation:
Sine (Periodicity), Trend, AR (Temporal dependence), Multi-freq
(Multi-periodicity), and Changepoint (Structural break).
The context length and the forecast horizon vary per sample, so no model can rely on a fixed window.
\begin{table}[t]
\centering
\setlength{\tabcolsep}{2pt}
\renewcommand{\arraystretch}{1.2}

\small
\begin{tabular}{@{}c|l|>{\centering\arraybackslash}m{2.4em}>{\centering\arraybackslash}m{2.1em}|>{\raggedright\arraybackslash}m{10.6em}@{}}
\toprule
 & \textbf{Pattern} & \textbf{Train} & \textbf{Test} & \textbf{Generative equation} \\
\midrule
\multirow{5}{*}{\rotatebox{90}{\textbf{Primitives}}}
  & Periodicity        & \multirow{5}{*}{O} & \multirow{5}{*}{O} & $s_1(t) + \varepsilon$ \\
  & Trend              &                &                & $s_1(t) + m t + b + \varepsilon$ \\
  & Temporal dependence&                &                & $\alpha\, x(t{-}1) + \varepsilon$ \\
  & Multi-periodicity  &                &                & $s_1(t) + s_2(t) + \varepsilon$ \\
  & Structural break   &                &                & Piecewise linear at~$t^{*}$ \\
\midrule
\multicolumn{2}{@{}>{\raggedright\arraybackslash}m{9.8em}|}{\shortstack[l]{\textbf{Unknown}\\\textit{(No primitive structure)}}}
    & O & O & Random walk, White noise, Exponential,
                Sawtooth, Square wave \\
\midrule
\multicolumn{2}{@{}>{\raggedright\arraybackslash}m{9.8em}|}{\shortstack[l]{\textbf{OOD-novel}\\\textit{(Variants of primitives)}}}
    & X & O & ARMA(1,1), Bilinear, Double-AR1,
                Quadratic
                \\
\bottomrule
\end{tabular}
\caption{
Overview of ReasonTS-Bench.
Five primitives and two out-of-pattern sets, where $s_i(t)=A_i\sin(2\pi t/P_i+\phi_i)$ is a periodicity term with amplitude $A_i$ and period $P_i$.}

\label{tab:bench}
\end{table}
To evaluate generalization \textit{beyond} the five primitives, we introduce two additional sets: Unknown and OOD-novel.
The \textbf{Unknown} set appears in both \textit{training} and \textit{testing}. 
For these inputs the model should report low confidence, state that no predefined pattern fits, and extrapolate the recent local trend. This encourages the model to avoid forcing a known pattern onto unfamiliar inputs %
(i.e., saying \textit{"I don’t know"}).
The \textbf{OOD-novel} set is used only for \textit{testing}. Its four series all differ from the
primitives but vary in how closely they resemble a primitive.
The members of each set are listed in Table~\ref{tab:bench}, 
with examples shown in Appendix~\ref{sec:appendix}.

\textbf{Details of datasets.} Appendix~\ref{sec:appendix} reports the per-split sample counts, the sequence lengths, and the full generative parameter ranges for every pattern in the benchmark.

\subsection{Reasoning Chain}

Every sample is paired with a reasoning chain in 
three blocks:

\begin{center}
\fbox{\parbox{0.90\columnwidth}{\ttfamily
\fontsize{9}{11}\selectfont
\textbf{INPUT ANALYSIS:}\\
\hspace*{1em}observed length, detected period, \dots\\
\textbf{REASONING:}\\
\hspace*{1em}pattern type, rule, next peak at, \dots\\
\textbf{PREDICTION:}\\
\hspace*{1em}t=N: $v_0$, t=N+1: $v_1$, \dots
}}
\end{center}

\begin{figure*}[t]
\centering
\adjustbox{max width=0.995\textwidth}{\includegraphics{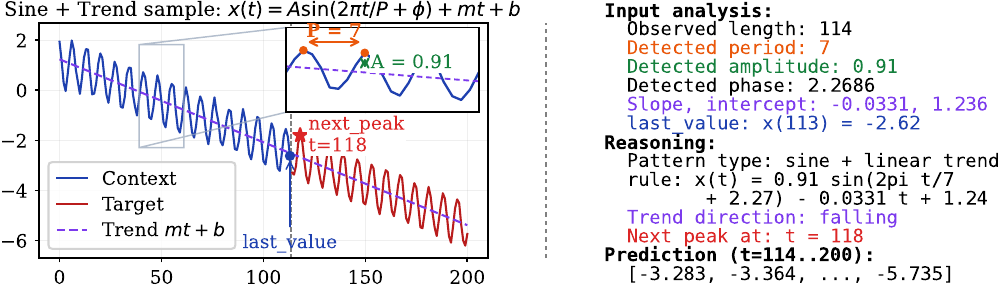}}
\caption{
Example of ground-truth reasoning.
[Left]
Sample with the trend line, the context and target split, and the next peak
and last value.
[Right]
Its ground-truth chain, where the color of each field marks the part of the series it is derived from.
}
\label{fig:worked}
\end{figure*}

Each named field on the right side of a colon is filled with a value derived
directly from the sample's generative parameters (e.g., detected period
is exactly $P$). This makes the chain verifiable, since an evaluator can parse
each field from the 
model output and compare it against the ground-truth target, using appropriate tolerances for continuous quantities.
Figure~\ref{fig:worked} provides a complete example for a Trend sample, showing the full ground-truth chain and how each field is derived from the series.
The schema is pattern-specific, and the fields required for each pattern are listed in Appendix~\ref{sec:appendix}.

\begin{table*}[t]
\centering
\setlength{\tabcolsep}{2pt}
\renewcommand{\arraystretch}{1.25}

\small
\begin{tabular}{c|cl|>{\centering\arraybackslash}p{0.35\textwidth}|>{\centering\arraybackslash}p{0.20\textwidth}|>{\centering\arraybackslash}p{0.14\textwidth}}
\toprule
 & & \textbf{Metric} & \textbf{[1] Summary} & \textbf{[2] What it compares} & \textbf{[3] Reduces to} \\
\midrule
\textbf{Forecasting} & (1) & \textbf{Error} ($\downarrow$) & Do the \textit{predicted numbers match the truth?} & $\hat{y}$ vs.\ $y$ & Regression \\
\midrule
\multirow{3}{*}{\textbf{Reasoning}} & (2) & \textbf{Fidelity} ($\uparrow$) & Are the \textit{stated facts grounded in the input?} & $\hat{\theta}$ vs.\ $\theta$ & Classification \\
 & (3) & \textbf{Consistency} ($\uparrow$) & Do the \textit{predictions obey the model's own rule?} & $\hat{y}$ vs.\ $\tilde{y}$ & Self-consistency \\
 & (4) & \textbf{Sensitivity} ($\uparrow$) & Does the \textit{reasoning follow a change in the input?} & $\hat{\theta}$ vs.\ $\hat{\theta}'$ under $\theta_j\!\to\!\theta_j'$ & Counterfactual \\
\bottomrule
\end{tabular}
\caption{Four metrics for TS forecasting and reasoning. Here $y$, $\hat{y}$, and $\tilde{y}=f_{\hat{\theta}}(x)$ denote the true value, the forecast generated directly by the LLM, and the value obtained by evaluating the estimated parameters $\hat{\theta}$ on the input $x$, respectively. $\theta$ and $\hat{\theta}$ are the true and estimated parameters. $\theta_j\!\to\!\theta_j'$ denotes a one-parameter intervention with re-estimated $\hat{\theta}'$.}
\label{tab:metrics}

\centering
\adjustbox{max width=\textwidth}{\includegraphics{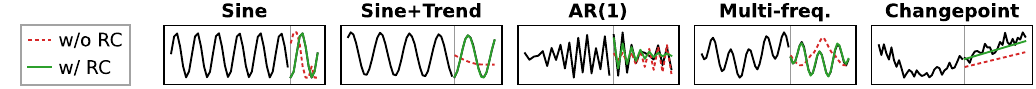}}
\captionof{figure}{Forecast visualization across the five primitives on the default
Qwen2.5-3B backbone, with and without ReasonCast. For each primitive one test
sample is shown: w/ ReasonCast tracks the ground truth, whereas w/o ReasonCast
does not.}
\label{fig:forecast5}
\end{table*}

\section{ReasonCast}
\label{sec:method}

ReasonCast is a \textit{recipe} that fine-tunes a \textit{single} model to route, reason, and forecast in \textit{one} autoregressive pass. It realizes the forward factorization of
Eq.~\eqref{eq:fwdfact},
generating the reasoning first
and the forecast second. 
Thus, the explanation \textit{precedes} and grounds the prediction, unlike prior work where the explanation \textit{follows} a given series.
The pipeline proceeds as follows:

\setlist[itemize]{leftmargin=8pt,itemsep=1pt,topsep=1pt}
\begin{itemize}
\item \textbf{Step 1. Read.} 
TS
is serialized into
the input, and the model conditions on it for everything that follows.
\item \textbf{Step 2. Route.} The model selects which output schema to emit,
that is, which pattern the input follows. 
Routing can be made 2-1) \textit{explicit} by
emitting the pattern class as the first token, or left 2-2) \textit{implicit} inside the
decoder.
\item \textbf{Step 3. Reason: $p_\theta(o_{\mathrm{rsn}} \mid X)$.} Conditioned on the selected pattern, the model
emits the 
reasoning chain, which states the estimated parameters and the
rule they imply.
\item \textbf{Step 4. Forecast: $p_\theta(o_{\mathrm{gen}} \mid o_{\mathrm{rsn}}, X)$.} Conditioned on the reasoning, the model emits
the numeric prediction that the rule implies, keeping the two mutually
consistent.
\item \textbf{Step 5. (Optional) Fall back.} 
If the input matches no
pattern, 
the model 
marks the pattern as unknown
and extrapolates the recent 
trend instead of forcing a pattern fit.
\end{itemize}

We study three ReasonCast variants.
1) \textbf{Explicit routing}, our default, emits the pattern class as its first
token, so routing is explicit and conditions what follows.
2) \textbf{Implicit routing} drops that token and selects the schema inside the decoder.
3) The \textbf{Single-pattern specialist} trains one model per pattern, with the pattern
given in advance, so it acts as an oracle.
Both routing variants outperform even the Single-pattern specialists, demonstrating that the five primitives are complementary and benefit from joint learning, as shown in Table~\ref{tab:routing} and Table~\ref{tab:complementary}.
Details are discussed in Appendix~\ref{sec:appendix_pipeline}.

\textbf{Evaluation metrics.}\label{sec:metrics} Conventional forecasting metrics
score the numbers alone, not whether they agree with the reasoning behind
them. We report four metrics, summarized in Table~\ref{tab:metrics}: \textbf{Error} (deviation of the predictions from the truth),
\textbf{Fidelity} (correctness of the estimated parameters), \textbf{Consistency}
(agreement between the forecast and the model's own rule), and \textbf{Sensitivity}
(causal response to a parameter change). Appendix~\ref{sec:appendix_metricdefs} defines each of them precisely.

\begin{table*}[t]
\centering\setlength{\tabcolsep}{2pt}\renewcommand{\arraystretch}{1.10}
\small
\begin{tabular}{@{}ll|cccccccccc|cc@{}}
\toprule
\multicolumn{2}{@{}l|}{\multirow{2.5}{*}{\textbf{Five primitives}}} & \multicolumn{2}{c}{\textbf{Sine}} & \multicolumn{2}{c}{\textbf{Trend}} & \multicolumn{2}{c}{\textbf{AR}} & \multicolumn{2}{c}{\textbf{MF}} & \multicolumn{2}{c|}{\textbf{CP}} & \multicolumn{2}{c}{\textbf{Average}} \\
\cmidrule(lr){3-4}\cmidrule(lr){5-6}\cmidrule(lr){7-8}\cmidrule(lr){9-10}\cmidrule(lr){11-12}\cmidrule(lr){13-14}
 &  & \textbf{MAE} & \textbf{MSE} & \textbf{MAE} & \textbf{MSE} & \textbf{MAE} & \textbf{MSE} & \textbf{MAE} & \textbf{MSE} & \textbf{MAE} & \textbf{MSE} & \textbf{MAE} & \textbf{MSE} \\
\midrule
\multicolumn{14}{@{}l}{\textbf{Axis 2: G}} \\
\midrule
\multirow{2}{*}{Trivial} & Naive last & 2.251 & 9.331 & 2.466 & 10.647 & 0.252 & 0.134 & 2.075 & 6.949 & 1.453 & 4.368 & 1.699 & 6.286 \\
 & Seasonal naive & 1.211 & 4.770 & 1.759 & 6.568 & 0.249 & 0.121 & 1.279 & 3.418 & 1.495 & 4.542 & 1.199 & 3.884 \\
\midrule
\multirow{4}{*}{TS models} & DLinear & 1.032 & 2.506 & 1.208 & 3.118 & 0.187 & 0.074 & 0.900 & 1.592 & 0.226 & 0.140 & 0.710 & 1.486 \\
 & PatchTST & 0.214 & 0.126 & 0.470 & 0.554 & 0.185 & 0.072 & 0.354 & 0.277 & 0.232 & 0.106 & 0.291 & 0.227 \\
 & iTransformer & 0.250 & 0.202 & 0.547 & 0.800 & 0.185 & 0.072 & 0.276 & 0.176 & 0.242 & 0.117 & 0.300 & 0.273 \\
 & TimeXer & 0.514 & 0.914 & 0.933 & 1.983 & 0.182 & 0.070 & 0.674 & 0.921 & 0.246 & 0.119 & 0.510 & 0.801 \\
\midrule
\multicolumn{14}{@{}l}{\textbf{Axis 3: U$+$G} (w/o ReasonCast),\quad \textbf{Axis 4: U$\times$G} (w/ ReasonCast)} \\
\midrule
\multirow{2}{*}{Qwen2.5-1.5B} & w/o ReasonCast & 2.507 & 12.628 & 3.735 & 27.977 & 0.189 & 0.080 & 2.169 & 8.257 & 2.021 & 12.135 & 2.124 & 12.215 \\
 & w/ ReasonCast & \first{0.203} & \first{0.108} & \first{0.393} & \first{0.364} & \first{0.183} & \first{0.076} & \first{0.345} & \first{0.431} & \first{0.188} & \first{0.141} & \first{0.262} & \first{0.224} \\
\midrule
\multirow{2}{*}{Qwen2.5-7B} & w/o ReasonCast & 2.185 & 8.768 & 3.085 & 18.497 & 0.196 & 0.091 & 2.294 & 9.701 & 0.846 & 1.877 & 1.721 & 7.787 \\
 & w/ ReasonCast & \first{0.229} & \first{0.174} & \first{0.471} & \first{0.639} & \first{0.181} & \first{0.073} & \first{0.520} & \first{0.807} & \first{0.180} & \first{0.104} & \first{0.316} & \first{0.359} \\
\midrule
\multirow{2}{*}{Llama-3.1-8B} & w/o ReasonCast & 2.616 & 13.345 & 4.426 & 36.476 & 0.189 & 0.088 & 4.373 & 34.206 & 1.878 & 8.208 & 2.696 & 18.465 \\
 & w/ ReasonCast & \first{0.436} & \first{0.702} & \first{0.809} & \first{1.515} & \first{0.186} & \first{0.091} & \first{0.694} & \first{1.211} & \first{0.242} & \first{0.169} & \first{0.474} & \first{0.737} \\
\midrule
\multirow{2}{*}{Phi-3.5-mini} & w/o ReasonCast & 2.531 & 12.067 & 3.899 & 27.209 & 0.245 & 0.158 & 2.309 & 9.536 & 2.304 & 11.847 & 2.258 & 12.163 \\
 & w/ ReasonCast & \first{0.542} & \first{0.981} & \first{1.312} & \first{4.292} & \first{0.182} & \first{0.074} & \first{1.061} & \first{2.538} & \first{0.245} & \first{0.277} & \first{0.668} & \first{1.632} \\
\midrule
\multirow{2}{*}{Gemma-2-9B} & w/o ReasonCast & 2.038 & 8.217 & 3.169 & 20.204 & 0.192 & 0.085 & 1.948 & 6.493 & 0.963 & 2.024 & 1.662 & 7.405 \\
 & w/ ReasonCast & \first{0.632} & \first{1.188} & \first{1.066} & \first{2.758} & \first{0.182} & \first{0.076} & \first{1.294} & \first{3.425} & \first{0.311} & \first{0.523} & \first{0.697} & \first{1.594} \\
\midrule
\multirow{2}{*}{TimeOmni-1} & w/o ReasonCast & 2.187 & 9.304 & 2.913 & 15.689 & 0.185 & 0.074 & 2.093 & 7.580 & 0.665 & 1.379 & 1.608 & 6.805 \\
 & w/ ReasonCast & \first{0.232} & \first{0.166} & \first{0.453} & \first{0.528} & \first{0.180} & \first{0.070} & \first{0.489} & \first{0.678} & \first{0.177} & \first{0.111} & \first{0.306} & \first{0.311} \\
\midrule
\multirow{2}{*}{\shortstack{Qwen2.5-3B\\(Default)}} & w/o ReasonCast & 2.375 & 10.398 & 3.526 & 24.319 & 0.243 & 0.229 & 1.904 & 6.218 & 3.065 & 22.864 & 2.223 & 12.805 \\
 & w/ ReasonCast & \first{0.208} & \first{0.139} & \first{0.360} & \first{0.343} & \first{0.181} & \first{0.073} & \first{0.256} & \first{0.225} & \first{0.177} & \first{0.145} & \first{0.236} & \first{0.185} \\
\bottomrule
\end{tabular}
\caption{Forecast performance. The five columns are the primitives: Sine, Trend, AR (AR(1)), MF (Multi-freq), and CP (Changepoint). The Axis~3 rows give each backbone few-shot examples so that it follows both the U and G output formats.}
\label{tab:forecast}
\end{table*}

\section{Experiments}
\label{sec:exp}

ReasonCast is a \textit{recipe rather than a single model}. Unless otherwise
stated, we use Qwen2.5-3B-Instruct~\citep{Yang2024Qwen25} as the default backbone, fully fine-tuned on
the ReasonTS-Bench training set, and write ReasonCast-3B for this default
instance. When we apply ReasonCast to other backbones,
we
fully fine-tune backbones up to 3B and use LoRA~\citep{Hu2022LoRA} for larger ones.
Results are averaged over 3 random seeds, with standard deviations for the default
model reported in Appendix~\ref{sec:appendix_forecast_std}.
We discuss the experimental setup in Appendix~\ref{sec:appendix_eval_protocol} and the dataset statistics in Appendix~\ref{sec:appendix}.

\textbf{Baseline.} We compare against baselines that span the four
axes of Table~\ref{tab:axes}: trivial predictors, four TS forecasting models
(DLinear, PatchTST, iTransformer, TimeXer~\citep{Wang2024TimeXer}), the (3)-axis unified model
TimeOmni-1, and four instruction-tuned LLM backbones 
(Qwen~\citep{Yang2024Qwen25}, Llama~\citep{Grattafiori2024Llama3},
Phi~\citep{Abdin2024Phi3}, Gemma~\citep{Riviere2024Gemma2}),
each run both few-shot and fine-tuned. 
Note that the reasoning metrics cannot be evaluated on the TS forecasting models, 
as they output only numeric values.
Details are provided in Appendix~\ref{sec:appendix_baseline_catalog}.
\begin{table}[!t]
\centering
\setlength{\tabcolsep}{1pt}
\renewcommand{\arraystretch}{1.11}
\small
\begin{tabular}{@{}l l| c c c c@{}}
\toprule
\multicolumn{2}{@{}l|}{\textbf{Avg.\ across 5 primitives}} & \textbf{Error}~$\downarrow$ & \textbf{Fid.}~$\uparrow$ & \textbf{Cons.}~$\uparrow$ & \textbf{Sens.}~$\uparrow$ \\
\midrule
\multirow{2}{*}{Qwen2.5-7B} & w/o ReasonCast & 1.721 & 0.338 & 0.250 & 0.142 \\
 & w/ ReasonCast & \first{0.316} & \first{0.859} & \first{0.532} & \first{0.714} \\
\midrule
\multirow{2}{*}{Llama-3.1-8B} & w/o ReasonCast & 2.696 & 0.293 & 0.240 & 0.138 \\
 & w/ ReasonCast & \first{0.474} & \first{0.708} & \first{0.341} & \first{0.574} \\
\midrule
\multirow{2}{*}{Phi-3.5-mini} & w/o ReasonCast & 2.258 & 0.262 & 0.218 & 0.154 \\
 & w/ ReasonCast & \first{0.668} & \first{0.690} & \first{0.332} & \first{0.542} \\
\midrule
\multirow{2}{*}{Gemma-2-9B} & w/o ReasonCast & 1.662 & 0.348 & 0.253 & 0.122 \\
 & w/ ReasonCast & \first{0.697} & \first{0.717} & \first{0.329} & \first{0.574} \\
\midrule
\multirow{2}{*}{\shortstack{Qwen2.5-3B\\(Default)}} & w/o ReasonCast & 2.223 & 0.252 & 0.275 & 0.138 \\
 & w/ ReasonCast & \first{0.236} & \first{0.899} & \first{0.613} & \first{0.794} \\
\bottomrule
\end{tabular}
\caption{Reasoning abilities. 
For each LLM backbone, we 
compare the effect of ReasonCast
on the forecast error (MAE)
and three reasoning metrics, averaged over five primitives.}

\label{tab:reasoning_compare}
\end{table}

\subsection{Main Experiments}
\textbf{Forecast performance (Error $\downarrow$).}
Table~\ref{tab:forecast} reports forecasting performance across the five primitives.
The results show that applying ReasonCast to a range of LLM backbones gives them a
TS forecasting ability that outperforms even the specialized TS forecasting models.
For a fair comparison, each of these LLMs is given a few
in-context examples so that it follows both the understanding and generation
output formats.
Figure~\ref{fig:forecast5} shows this across all five primitives: w/ ReasonCast tracks the truth while w/o ReasonCast quickly drifts off.

\begin{table}[t]
\centering

\centering
\setlength{\tabcolsep}{2pt}
\renewcommand{\arraystretch}{1.15}
\small
\begin{tabular}{@{}l| c c c@{}}
\toprule
\textbf{Output order} & \textbf{Error}~$\downarrow$ & \textbf{Fidelity}~$\uparrow$ & \textbf{Consistency}~$\uparrow$ \\
\midrule
Forecast $\to$ Reasoning & 0.471 & 0.293 & 0.345 \\
Reasoning $\to$ Forecast & \textbf{0.404} & \textbf{0.837} & \textbf{0.465} \\
\midrule
$\Delta$ (improvement) & $-0.070$ & $+55$\,pp & $+13$\,pp \\
\bottomrule
\end{tabular}
\caption{Order ablation: reasoning before forecast. Reasoning first grounds the forecast and keeps the explanation faithful.}
\label{tab:order}

\end{table}

\textbf{Reasoning abilities (Fidelity $\uparrow$, Consistency $\uparrow$, Sensitivity $\uparrow$).}
Table~\ref{tab:reasoning_compare} compares each backbone with and without ReasonCast on the forecast error (MAE) and three reasoning metrics, averaged over five primitives.
The results demonstrate that ReasonCast lowers forecast error while raising every reasoning metric, yielding forecasts that come with an explanation of why each value was predicted. 

\subsection{Ablation Studies}

\textbf{Order of reasoning \& forecast.}
To verify that \textit{the reasoning genuinely grounds the forecast} rather than merely accompanying
it, we train a post-hoc variant that emits the \textit{forecast first and the reasoning
afterward}, changing only the output order. As shown in Table~\ref{tab:order}, this
reversed ordering consistently degrades performance relative to reasoning first,
demonstrating that reasoning steers the forecast rather than merely explaining it.
We omit Sensitivity as Fidelity already isolates it.

\begin{figure}[t]
\centering
\adjustbox{width=\columnwidth}{\includegraphics{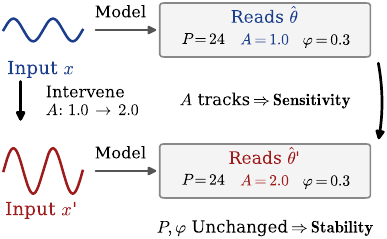}}
\caption{How the counterfactual probe is scored. A paired input differing
in a single generative parameter is fed to the model, which should shift the
intervened field (Sensitivity) while leaving the others unchanged (Stability).}
\label{fig:cf_schematic}
\end{figure}

\begin{table}[t]\centering\setlength{\tabcolsep}{2pt}\renewcommand{\arraystretch}{1.15}
\small
\begin{tabular}{@{}l l| c c c c c@{}}
\toprule
 & & \textbf{Sine} & \textbf{Trend} & \textbf{AR} & \textbf{MF} & \textbf{CP} \\
\midrule
\multirow{2}{*}{\textbf{Sensitivity}~$\uparrow$} & w/o ReasonCast & 0.270 & 0.200 & 0.050 & 0.070 & 0.100 \\
 & w/ ReasonCast & \first{0.960} & \first{0.880} & \first{0.530} & \first{0.820} & \first{0.780} \\
\midrule
\multirow{2}{*}{\textbf{Stability}~$\uparrow$} & w/o ReasonCast & 0.705 & 0.690 & -- & 0.680 & 0.650 \\
 & w/ ReasonCast & \first{0.905} & \first{0.890} & -- & \first{0.765} & \first{0.840} \\
\bottomrule
\end{tabular}
\caption{Counterfactual probe, w/ vs.\ w/o ReasonCast. ``Sensitivity'' is the intervention-tracking rate (the matching reasoning field shifts to track an intervened parameter), and ``Stability'' is the non-intervened-stable rate (the remaining reasoning fields stay fixed under the same intervention).}
\label{tab:cf}
\end{table}

\textbf{Counterfactual probing.}
The counterfactual probe tests \textit{whether 
the model reads the input causally}
or merely reproduces memorized training facts. 
Each test item is a pair $(x, x')$ identical except for one intervened generative parameter
$\theta_j \!\to\! \theta_j'$ (e.g., amplitude of Sine). 
As shown in Figure~\ref{fig:cf_schematic}, we run the model on both,
parse the reasoning fields it states for each, and check two things: (1) \textit{Sensitivity}, whether
the matching field changes with the intervention, and (2) \textit{Stability}, whether the untouched
fields stay unchanged.
A causal model changes only the intervened field and keeps the untouched fields unchanged, whereas a memorizer changes nothing despite the intervention. 
Details of the Sensitivity and Stability computations are provided in Appendix~\ref{sec:appendix_cfprobe}.

Table~\ref{tab:cf} reports both rates on each pattern. Without ReasonCast the few-shot model rarely tracks the change, giving low Sensitivity and only moderate Stability. ReasonCast raises Sensitivity sharply on every pattern and lifts Stability as well, showing that the trained model reads the input rather than recalling a memorized value. Appendix~\ref{sec:appendix_cfprobe} shows one tracked and one ignored example. As AR has only one field, no other field remains to check, and Stability is undefined.
\begin{table}[t]
\centering\setlength{\tabcolsep}{3pt}\renewcommand{\arraystretch}{1.12}
\small
\begin{tabular}{@{}l | c c c c c | c@{}}
\toprule
\textbf{Model} & \textbf{Sine} & \textbf{Trend} & \textbf{AR} & \textbf{MF} & \textbf{CP} & \textbf{Average} \\
\midrule
w/o ReasonCast & 0.153 & 0.165 & 0.440 & 0.164 & 0.341 & 0.252 \\
\midrule
\multicolumn{7}{@{}l}{w/ ReasonCast} \\
Implicit routing & 0.924 & 0.888 & \first{0.870} & 0.868 & 0.921 & 0.894 \\
Explicit routing & \first{0.926} & \first{0.898} & 0.864 & \first{0.883} & \first{0.924} & \first{0.899} \\
\bottomrule
\end{tabular}
\caption{Robustness to explicit vs.\ implicit routing: either variant lifts Fidelity far above w/o ReasonCast, and the two variants land at essentially the same accuracy.}
\label{tab:routing}

\centering\setlength{\tabcolsep}{2pt}\renewcommand{\arraystretch}{1.12}
\small
\begin{tabular}{@{}l | c c c c c | c@{}}
\toprule
\textbf{Model} & \textbf{Sine} & \textbf{Trend} & \textbf{AR} & \textbf{MF} & \textbf{CP} & \textbf{Average} \\
\midrule
Separate (5 Models) & 0.766 & 0.827 & \textbf{0.864} & 0.764 & 0.920 & 0.827 \\
Joint (1 Model) & \textbf{0.926} & \textbf{0.898} & \textbf{0.864} & \textbf{0.883} & \textbf{0.924} & \textbf{0.899} \\
\bottomrule
\end{tabular}
\caption{Separate vs.\ Joint training, where a single joint model outperforms the five separate specialists, showing that the primitives are complementary.}
\label{tab:complementary}

\end{table}

\textbf{Robustness to explicit and implicit routing.} To show that the gain does not
depend on how the pattern is routed, we evaluate Explicit and Implicit routing
against w/o ReasonCast. In Table~\ref{tab:routing},
both routing variants raise average Fidelity by a wide margin, and ReasonCast is
robust to how the pattern is routed. This demonstrates that the gain comes from the
reasoning itself, not from how the pattern class is exposed.

\section{Analysis}
\label{sec:analysis}
In this section, we analyze the following six aspects:
\begin{itemize}[leftmargin=1.2em, itemsep=1pt, topsep=2pt, parsep=0pt]
\item Complementarity of the five primitives
\item Grounding from reasoning content, not format
\item Performance on Unknown and OOD-novel inputs
\item Application to real-world data
\item Comparison with symbolic regression
\item Capacity scaling (Appendix~\ref{sec:appendix_scaling})
\end{itemize}

\begin{figure*}[t]
\centering
\adjustbox{max width=\textwidth}{\includegraphics{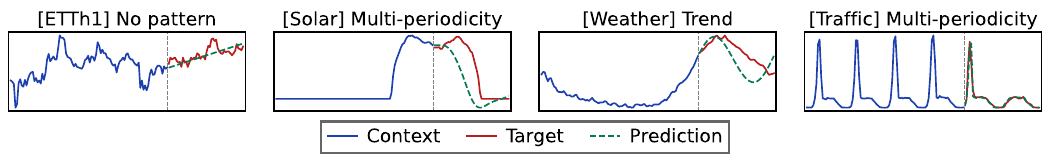}}
\caption{Application to real-world datasets: ETTh1, Solar, Weather, Traffic. The
default model is applied without further training, with the pattern class the model
routed each window to. Each window has a context of 200 steps and a horizon of 48.}
\label{fig:realworld}
\end{figure*}

\textbf{Complementarity of the five primitives.} Table~\ref{tab:complementary}
compares the routed single model against the Single-pattern specialists.
Explicit routing reaches a higher accuracy than the specialists that each
see a single pattern. The results demonstrate that the five primitives are complementary, and one model
can replace the five separate specialists at deployment.

\textbf{Grounding from reasoning content, not format.} While Table~\ref{tab:order} shows that emitting the reasoning first matters, it does not say whether the grounding comes from the fields the model learns to emit or from the values it puts in them. To test whether the reasoning is genuinely grounded in the input rather than only following the required format, we \textit{replace each sample's reasoning target} with the reasoning of another sample. The forecast target stays correct, so the reasoning is the only part of the supervision that no longer describes the input.
As shown in Table~\ref{tab:corrupt}, the corrupted model degrades on every pattern even though its forecast targets were left untouched. This demonstrates that the \textit{reasoning and the forecast depend on each other}, 
and that correct fields alone do not make either one right,
as Appendix~\ref{sec:appendix_grounding} illustrates.

\begin{table}[t]
\centering
\setlength{\tabcolsep}{2pt}
\renewcommand{\arraystretch}{1.15}
\small
\begin{tabular}{@{}l | ccc | ccc@{}}
\toprule
 & \multicolumn{3}{c|}{\textbf{Corrupted reasoning}} & \multicolumn{3}{c}{\textbf{Correct reasoning}} \\
\cmidrule(lr){2-4}\cmidrule(lr){5-7}
\textbf{Pattern} & \textbf{Fid.}~$\uparrow$ & \textbf{Cons.}~$\uparrow$ & \textbf{Error}~$\downarrow$ & \textbf{Fid.}~$\uparrow$ & \textbf{Cons.}~$\uparrow$ & \textbf{Error}~$\downarrow$ \\
\midrule
Sine & 0.055 & 0.073 & 0.558 & \textbf{0.766} & \textbf{0.383} & \textbf{0.549} \\
Trend & 0.139 & 0.032 & 1.136 & \textbf{0.827} & \textbf{0.219} & \textbf{0.647} \\
AR(1) & 0.386 & 0.963 & 0.179 & \textbf{0.884} & \textbf{1.000} & \textbf{0.176} \\
Multi-freq & 0.000$^{\dagger}$ & -- & -- & \textbf{0.764} & \textbf{0.189} & \textbf{0.454} \\
Changepoint & 0.258 & 0.009 & 0.203 & \textbf{0.944} & \textbf{0.534} & \textbf{0.195} \\
\midrule
Average & 0.167 & 0.269 & 0.505 & \textbf{0.837} & \textbf{0.465} & \textbf{0.404} \\
\bottomrule
\end{tabular}
\caption{Reasoning content ablation. Swapping each training reasoning for another sample's makes fields wrong but the prediction correct. $^{\dagger}$ Multi-freq emitted nothing parseable.}
\label{tab:corrupt}
\end{table}

\textbf{Application to real-world data.} We also apply the default model, with no
further training, to windows drawn from the four real-world datasets of
Figure~\ref{fig:coverage}. As Figure~\ref{fig:realworld} shows, the model routes each
window either to one of the five patterns or to no-pattern, and it forecasts that
window under the choice it made. The pattern it settles on differs across datasets,
with Multi-periodicity on Solar and Traffic and Trend on Weather. On the ETTh1 window
it reports no pattern, yet its forecast still follows the target by extrapolating the
recent trend.

\textbf{Performance on Unknown and OOD-novel inputs.} We also test the two sets of
inputs in Table~\ref{tab:bench} that match none of the five primitives. As
Table~\ref{tab:ood} shows, ReasonCast forecasts them more accurately than
w/o ReasonCast, both on the unknown inputs it was trained on and on the
held-out processes it has never seen. The gain holds for either routing variant,
which indicates that it comes from the reasoning rather than from how the model
routes.
\begin{table}[t]\centering\setlength{\tabcolsep}{2pt}\renewcommand{\arraystretch}{1.12}
\small
\begin{tabular}{@{}l | c | c c@{}}
\toprule
 & \multirow{2.5}{*}{\textbf{w/o ReasonCast}} & \multicolumn{2}{c}{\textbf{w/ ReasonCast}} \\
\cmidrule(lr){3-4}
\textbf{No-pattern} & & \textbf{Implicit} & \textbf{Explicit} \\
\midrule
\multicolumn{4}{@{}l}{\textit{Unknown (in-distribution)}} \\
Random walk & 1.70 & \first{1.34} & \first{1.39} \\
White noise & 2.02 & \first{1.51} & \first{1.52} \\
Exponential & 187.86 & \first{108.4} & \first{108.9} \\
Sawtooth & 1.04 & \first{0.82} & \first{0.80} \\
Square wave & 6.26 & \first{4.60} & \first{4.65} \\
\midrule
\multicolumn{4}{@{}l}{\textit{OOD-novel (held out)}} \\
Quadratic trend & 24.34 & \first{4.41} & \first{6.42} \\
ARMA(1,1) & 0.75 & \first{0.49} & \first{0.50} \\
Double AR(1) & 1.00 & \first{0.65} & \first{0.67} \\
Bilinear & 1.62 & \first{1.16} & \first{1.61} \\
\bottomrule
\end{tabular}
\caption{Forecast error on inputs with no pattern. Each entry is the average MAE over the test samples of that process. Bold marks a gain over w/o ReasonCast.}
\label{tab:ood}
\end{table}

\textbf{Symbolic regression: structure without explanation.} Symbolic regression~\citep{SchmidtLipson2009,SINDy2016} searches
a space of mathematical expressions for a closed-form equation that fits the observed
series. It selects the
expression whose parameters best reproduce the series, so the
fitted formula is at once the forecaster and its own explanation of the dynamics.
However, compared with ReasonCast, the fitted equation is all it offers. As
Table~\ref{tab:symbolic} shows, symbolic regression 1) gives no natural-language
account, 2) cannot flag unfamiliar inputs, 3) provides no forecast-consistency check,
and 4) forecasts less accurately.
Details are provided in Appendix~\ref{sec:appendix_symbolic}.

\begin{table}[t]%
\small
\begin{tabular}{@{}l|cccc|c}
\toprule
 & \multicolumn{4}{c|}{\textbf{Capabilities}} & \multirow{2.5}{*}{\textbf{Error}~$\downarrow$} \\
\cmidrule(lr){2-5}
\textbf{Method} & \textbf{Struct.} & \textbf{Lang.} & \textbf{Unfam.} & \textbf{Cons.} & \\
\midrule
Symbolic & \checkmark & -- & -- & -- & 0.320 \\
ReasonCast & \checkmark & \checkmark & \checkmark & \checkmark & \textbf{0.236} \\
\bottomrule
\end{tabular}
\caption{Symbolic regression vs.\ ReasonCast.}
\label{tab:symbolic}
\end{table}

\section{Conclusion}

We argue for a fourth axis of TS modeling, task-fused understanding $\times$
generation, in which a model predicts and self-explains in a single
output. We instantiate it with ReasonTS-Bench, the first TS benchmark
with step-verifiable reasoning chains and four metrics,
and with ReasonCast, a recipe that yields one deployable model
outperforming baselines while providing grounded,
no-pattern-capable reasoning.

\textbf{Limitations and future work.} ReasonTS-Bench is a synthetic
benchmark built on five primitives, and two extensions stand out. First,
it is univariate, and moving to multivariate series calls for explanations that must
also account for cross-variate dependence. Second, its five primitives are elementary
building blocks, and grounding explanations in their compositions is the
natural path toward richer explainability.

\bibliography{aaai2027}

\clearpage
\appendix
\section{Additional Related Work}
\label{sec:appendix_relatedwork}

\textbf{TS benchmarks.} Table~\ref{tab:related_all} also groups TS
benchmarks by the four axes. Forecasting benchmarks such as the Monash
archive~\citep{Monash}, GIFT-Eval~\citep{GIFTEval}, and TFB~\citep{TFB} score
numeric accuracy alone (axis G). A newer line evaluates TS reasoning in language, where
the TSQA set of Time-MQA~\citep{TimeMQA}, MTBench~\citep{MTBench}, and
SenTSR-Bench~\citep{SenTSRBench} pose textual questions about a series (axis U).
Closest to us, the TSR-Suite of TimeOmni-1~\citep{TimeOmni1} and
FinSTaR~\citep{FinSTaR} pair reasoning with prediction, but as separate task
families (axis U+G). The idea of a step-verifiable reasoning chain follows
chain-of-thought benchmarks in math and code~\citep{GSM8K,MATH,BBH}, where each
step is checkable. To our knowledge, \textit{ReasonTS-Bench is the first to bring this to
TS (axis U$\times$G)}: it asks a model to derive a numeric forecast from observed
dynamics in a chain that is verifiable at the step level and scored \textit{jointly} with
the forecast.

\textbf{Symbolic regression.} The closest classical analog to a model that both predicts
and explains is symbolic regression, which recovers a closed-form expression that fits
observed data~\citep{SchmidtLipson2009,SINDy2016}. A fitted expression is at once a
predictor and an account of the data, so it unifies generation and understanding in a
single object. It differs from our setting in two ways. The explanation is a symbolic
formula rather than a natural-language justification, and the method emits one fused
expression rather than a forecast and a separate rationale whose mutual consistency can
be scored. We compare against a symbolic-regression baseline in the analysis.

\section{Metric Definitions}
\label{sec:appendix_metricdefs}

\begin{figure*}[t]
\centering
\adjustbox{max width=\textwidth}{\includegraphics{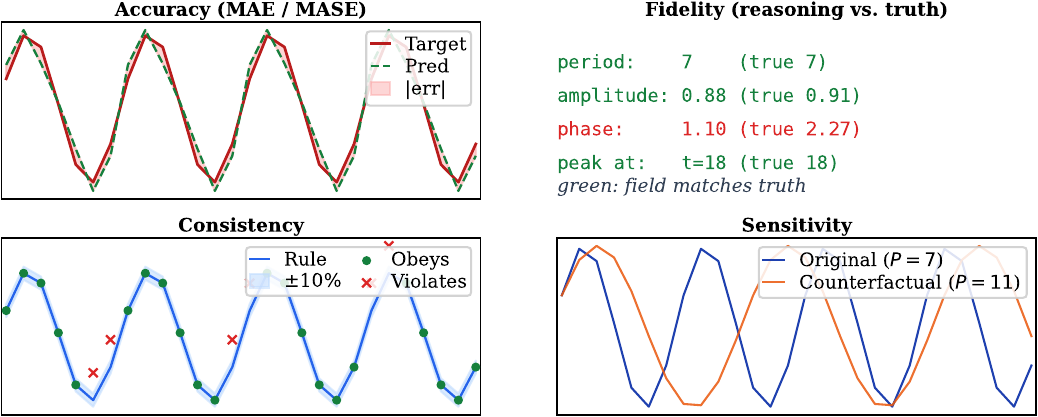}}
\caption{The four metric families, illustrated on a Sine example.
Error averages the shaded gap between prediction and target. Fidelity checks
each reasoning field against the closed-form truth (green: match, red: miss).
Consistency checks whether predicted points obey the model's stated rule (green)
or violate it (red $\times$), needing no ground truth. Sensitivity swaps one
parameter (period $7\!\to\!11$) and asks whether only the matching field moves.}
\label{fig:metrics}
\end{figure*}

This section gives the exact computation of the four metric families introduced
in Section~\ref{sec:metrics}. We write a sample as a context $x_{1:N}$ followed
by a target $y=(y_1,\dots,y_H)$ with $y_k=x_{N+k}$, and the model emits a
forecast $\hat{y}=(\hat{y}_1,\dots,\hat{y}_H)$ together with a reasoning chain
whose named fields we collect in a set $\mathcal{F}$. For a field $f$, $\hat{v}_f$
is the value the model emits and $v_f^{\star}$ is the closed-form ground truth
computed from the generative parameters. Figure~\ref{fig:metrics} works all four
families through a single Sine example, showing on the same series what each one
measures.

\begin{table*}[t]
\centering
\setlength{\tabcolsep}{2pt}
\renewcommand{\arraystretch}{1.18}

\small
\begin{tabular}{@{}l|ll@{}}
\toprule
\textbf{Pattern} & \textbf{Equation} & \textbf{Parameter ranges} \\
\midrule
Periodicity         & $A\sin(2\pi t/P + \phi) + \varepsilon$
                   & $P\in[4,50],\; A\in[0.5,5],\; \phi\in[0,2\pi]$\\
Trend    & $A\sin(2\pi t/P + \phi) + m t + b + \varepsilon$
                   & Periodicity $+$ $m\in[-0.05,0.05],\,b\in[-2,2]$\\
Temporal dependence & $\alpha\, x(t{-}1) + \varepsilon$
                   & $\alpha\in[-0.95,0.95],\,\sigma\in[0.05,0.3]$\\
Multi-periodicity   & $\sum_{i=1,2} A_i \sin(2\pi t/P_i + \phi_i) + \varepsilon$
                   & $P_1\in[3,20],\; P_2\geq 2P_1,\; P_2\leq 60$ \\
Structural break    & Piecewise linear at $t^{*}$
                   & $t^{*}\in[0.25N,0.85N],\,m_1,m_2\in[-0.1,0.1]$\\
\midrule
Unknown         & Random walk, white noise, exponential,
                     sawtooth, square wave
                   & \textit{Distinct from all primitives} \\
OOD-novel          & ARMA(1,1), bilinear, double-AR1,
                     quadratic trend
                   & \textit{Truly unseen, used only at test time} \\
\bottomrule
\end{tabular}
\caption{ReasonTS-Bench patterns with generative equations and full
parameter ranges. $N$ denotes context length. The noise term $\varepsilon$ has
zero mean and standard deviation in $[0, 0.1]\cdot A$ for sinusoidal patterns.}
\label{tab:levels_full}
\end{table*}

\textbf{Matching a single field.}
Discrete fields (a period, a changepoint index, a state label) require an exact
match, and continuous fields (an amplitude, a slope) pass within a tolerance. We
write this as a single indicator with per-field absolute and relative tolerances
$a_f$ and $r_f$,
\begin{equation}
\begin{split}
\mathrm{match}(\hat{v}_f, v_f^{\star}) = \mathbf{1}\Bigl[\,
&|\hat{v}_f - v_f^{\star}| \le a_f \ \ \text{or}\\[-1pt]
&\ \ |\hat{v}_f - v_f^{\star}| \le r_f\,|v_f^{\star}| \,\Bigr],
\end{split}
\end{equation}
where a discrete field sets $a_f=r_f=0$, so the condition reduces to the exact
equality $\hat{v}_f=v_f^{\star}$, while a field that the model fails to emit is
counted as a mismatch and receives the minimum score of $0$ from the same indicator.
Table~\ref{tab:tol} lists the tolerances for every field.

\subsection{Error}
Error is the standard forecast deviation on the predicted values, parsed from the
PREDICTION block and aligned to the target by absolute time index. We report mean
absolute error and its scaled version,
\begin{equation}
\mathrm{MAE} = \frac{1}{H}\sum_{k=1}^{H} \bigl|\hat{y}_k - y_k\bigr|,
\end{equation}
\begin{equation}
\mathrm{MASE} = \frac{(H-1)\,\mathrm{MAE}}{\sum_{i=2}^{H} |y_i - y_{i-1}|},
\end{equation}
where MASE normalizes by the average one-step change of the target, so a value
below $1$ beats the naive forecast.

\textbf{Worked example.}
Take a four-step target $y=(2.0, 2.5, 2.4, 1.8)$ and a forecast
$\hat{y}=(2.1, 2.3, 2.6, 1.9)$. The absolute errors are
$(0.1, 0.2, 0.2, 0.1)$, so $\mathrm{MAE}=0.6/4=0.15$. The naive one-step
differences of the target are $(0.5, 0.1, 0.6)$ with mean $0.4$, so
$\mathrm{MASE}=0.15/0.4=0.375$, a value below $1$ that beats the naive
last-value forecast.

\subsection{Fidelity}
Fidelity checks whether the reasoning fields are correct rather than
hallucinated. Each field is matched against its closed-form truth, and the score
is the fraction of fields that match,
\begin{equation}
\mathrm{Fidelity} = \frac{1}{|\mathcal{F}|}\sum_{f\in\mathcal{F}}
\mathrm{match}(\hat{v}_f, v_f^{\star}).
\end{equation}
Fidelity is then averaged over all samples in the split.

\textbf{Worked example.}
Suppose the truth is $\text{period}^{\star}=36$, $\text{amplitude}^{\star}=3.71$,
$\text{phase}^{\star}=1.94$. The model emits $\widehat{\text{period}}=36$,
$\widehat{\text{amplitude}}=3.65$, $\widehat{\text{phase}}=2.31$. The period is
discrete and matches exactly. The amplitude is continuous with $a_f=0.1$, and
$|3.65-3.71|=0.06\le 0.1$ matches. The phase has $a_f=0.3$, and
$|2.31-1.94|=0.37>0.3$ fails. Fidelity is then $(1+1+0)/3\approx 0.67$.

\subsection{Consistency}
Consistency checks whether the predicted numbers obey the rule the model itself
states, and it uses no ground truth. The reasoning fields $\hat{o}$ define a rule
$g_{\hat{o}}$, and at each step we compute the value that rule implies,
$\tilde{y}_k = g_{\hat{o}}(N+k)$. The rule is pattern-specific. For a sinusoid
$\tilde{y}_k=\hat{A}\sin\!\big(2\pi (N{+}k)/\hat{P}+\hat{\phi}\big)$, for an AR(1)
process $\tilde{y}_k\approx\hat{\alpha}^{\,k+1}\,x_{N-1}$, and analogously for the
other patterns. The score is the fraction of forecast points
that lie within a tolerance of their own implied value,
\begin{equation}
\mathrm{Consistency} = \frac{1}{H}\sum_{k=1}^{H}
\mathbf{1}\!\left[\,\bigl|\hat{y}_k - \tilde{y}_k\bigr| \le \tau \cdot A\,\right],
\end{equation}
with $\tau=0.1$ and $A$ a per-pattern amplitude scale. Because $y$ never enters,
this isolates self-grounding from forecast error.

\textbf{Worked example.}
Suppose the model states the rule $\hat{x}(t)=3.65\sin(2\pi t/36 + 1.94)$, which
implies $\tilde{y}=(1.35, 1.93, 2.45, 2.90)$ over four steps, while its
PREDICTION block emits $\hat{y}=(1.40, 1.88, 2.52, 2.84)$. The deviations
$(0.05, 0.05, 0.07, 0.06)$ all fall within $\tau A = 0.1\times 3.71 = 0.371$, so
all four points are consistent and the score is $4/4=1.0$, even though no ground
truth was used.

\subsection{Sensitivity}
Sensitivity is measured with paired counterfactual probes. From a base sample
with generative parameters $\theta$ we build a counterfactual that changes one
parameter $\theta_j\!\to\!\theta_j'$ and holds the rest fixed, 
so the only difference between the two is the intervened
parameter. We run the model on both and read the reasoning fields $\hat{v}$ and
$\hat{v}'$. Let $f(j)$ be the field that should track parameter $j$. We report
two rates,
\begin{align}
\mathrm{Sensitivity} &= \mathrm{match}\!\left(\hat{v}'_{f(j)},\, \theta_j'\right),\\
\mathrm{Stability} &= \frac{1}{|\mathcal{F}\setminus\{f(j)\}|}
\sum_{f\neq f(j)} \mathrm{match}\!\left(\hat{v}_f,\, \hat{v}'_f\right),
\end{align}
averaged over pairs. A causally sound model scores high on both, meaning the
intervened field moves to the new value while every other field stays fixed.
This separates causal reasoning from memorization. A
model that merely matched an input to its nearest training instance could score
high on Fidelity yet low on Sensitivity.

The intervention redraws the chosen parameter instead of adding a fixed offset,
and Table~\ref{tab:cf_spec} lists the parameters probed for each pattern. A pair
counts as sensitive only when the stated field tracks the new value.

\section{Full Benchmark Specification}
\label{sec:appendix}

Table~\ref{tab:levels_full} gives the generative parameter ranges for every
pattern, Table~\ref{tab:schema} lists the reasoning fields the model must emit,
Table~\ref{tab:bench_excerpt} shows a ground-truth reasoning excerpt per pattern,
and Table~\ref{tab:splits} gives the size of every split.

\begin{table*}[p]
\centering
\setlength{\tabcolsep}{2pt}
\renewcommand{\arraystretch}{1.15}

\small
\begin{tabular}{@{}cp{0.82\textwidth}@{}}
\toprule
\textbf{Pattern} & \textbf{Reasoning fields the model must emit} \\
\midrule
Periodicity & Detected period, detected amplitude, detected phase, last value, current state, next peak at, next trough at, next cycle completes at \\
\midrule
Trend & Periodicity fields $+$ detected slope, detected intercept, trend direction \\
\midrule
Temporal dependence & Detected alpha, detected noise std, regime, half life steps, forecast rule \\
\midrule
Multi-periodicity & Detected period 1, detected amplitude 1, detected phase 1, detected period 2, detected amplitude 2, detected phase 2, dominant component, amplitude ratio \\
\midrule
Structural break & Detected changepoint t, segment 1 slope, segment 1 intercept, segment 2 slope, segment 2 intercept, current segment, regime after changepoint \\
\midrule
Unknown & Pattern match confidence: low, pattern type: unknown, fallback strategy: naive last value continuation \\
\bottomrule
\end{tabular}
\caption{Schema per pattern. A multi-pattern model must decide which schema to
emit per input, and this routing problem motivates the Implicit routing and Explicit routing variants
of Section~\ref{sec:method}.}
\label{tab:schema}
\end{table*}

\begin{table*}[p]
\centering
\small
\setlength{\tabcolsep}{1pt}
\begin{tabular}{@{}c|l|l|l|l|l@{}}
\toprule
 & \textbf{Sine} & \textbf{Trend} & \textbf{AR(1)} & \textbf{Multi-freq} & \textbf{Changepoint} \\
\midrule
\multirow{6}{*}{\textbf{Reasoning}} & \texttt{Pattern: Sine} & \texttt{Pattern: Trend} & \texttt{Pattern: AR(1)} & \texttt{Pattern: 2-Freq} & \texttt{Pattern: Break} \\
 & \texttt{Period: 7} & \texttt{Period: 7} & \texttt{Alpha: 0.87} & \texttt{Fast P: 3} & \texttt{Changept: 67} \\
 & \texttt{Amp: 0.91} & \texttt{Slope: -0.033} & \texttt{Regime: Decay} & \texttt{Slow P: 59} & \texttt{Slope 1: 0.099} \\
 & \texttt{Phase: 2.27} & \texttt{Dir: Falling} & \texttt{a x(t-1)+e} & \texttt{Amps: 2.79/1.04} & \texttt{Slope 2: -0.095} \\
 & \texttt{A sin(t/P)} & \texttt{sin + mt+b} & \texttt{Next: revert to 0} & \texttt{sin1+sin2} & \texttt{break at t*} \\
 & \texttt{Next: continue} & \texttt{Next: trend holds} &  & \texttt{Sep: 19.7x} & \texttt{Next: new slope} \\
\midrule
\multirow{2}{*}{\textbf{Prediction}} & \texttt{H=42 (128-169)} & \texttt{H=87 (114-200)} & \texttt{H=56 (71-126)} & \texttt{H=16 (133-148)} & \texttt{H=92 (146-237)} \\
 & \texttt{[-.71,..,-.02]} & \texttt{[-3.3,..,-5.7]} & \texttt{[-.70,..,-.02]} & \texttt{[-.73,..,-1.0]} & \texttt{[.77,..,-7.1]} \\
\bottomrule
\end{tabular}

\caption{Ground-truth reasoning excerpt for each sample plotted in
Figure~\ref{fig:bench}, where every field is computed in closed form from that
sample's generative parameters. The full chains appear in Appendix~\ref{sec:appendix_examples}.}
\label{tab:bench_excerpt}
\end{table*}

\begin{figure*}[p]
\centering
\adjustbox{max width=\textwidth}{\includegraphics{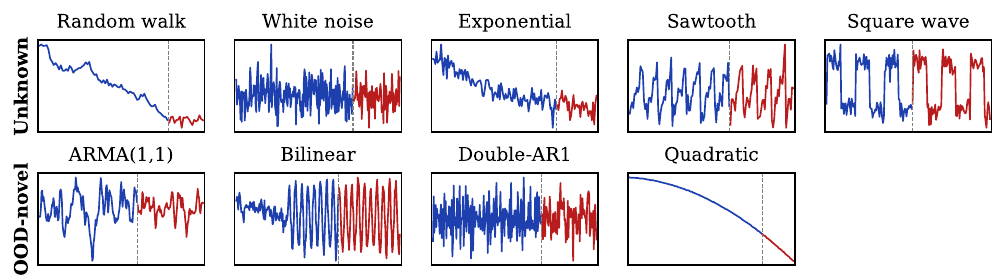}}
\caption{The two out-of-pattern sets. One sample is shown per member of the Unknown (top) and OOD-novel (bottom) sets, where blue marks the input context and red marks the forecast target that follows it.}
\label{fig:nonprim}
\end{figure*}

\begin{table*}[p]
\centering
\setlength{\tabcolsep}{2pt}
\renewcommand{\arraystretch}{1.15}
\small
\begin{tabular}{@{}l|ccc|c@{}}
\toprule
\textbf{Split} & \textbf{Train} & \textbf{Val} & \textbf{Test} & \textbf{Use} \\
\midrule
Primitives (each of 5) & 12{,}000 & 1{,}000 & 1{,}000 & Train and evaluate the model on each of the five primitive patterns \\
Unknown                & 12{,}000 & 1{,}000 & 1{,}000 & Learn to flag an unfamiliar, out-of-distribution input as no-pattern \\
OOD-novel                 & --       & --       & 500     & Test whether the no-pattern flagging transfers to unseen, held-out processes \\
CF probe (paired)         & --       & --       & 100--150/pattern & Paired inputs differing in a single generative parameter, for the causal check \\
\bottomrule
\end{tabular}
\caption{Dataset sizes. Context length $N\!\in\![50,200]$ and horizon $H\!\in\![10,100]$ vary per sample.}
\label{tab:splits}
\end{table*}

\textbf{The two out-of-pattern sets.}
\label{sec:appendix_nonprim}
Figure~\ref{fig:nonprim} shows one sample from each member of the two sets.
Unknown holds five signals that match no primitive and differ from one another,
which teaches the model to flag unfamiliar inputs rather than memorize one shape.
OOD-novel holds four processes never seen in training, ordered from far from any
primitive to close to one, which lets us measure how far the flagging carries.

\begin{figure*}[t]
\centering
\adjustbox{max width=\textwidth}{\includegraphics{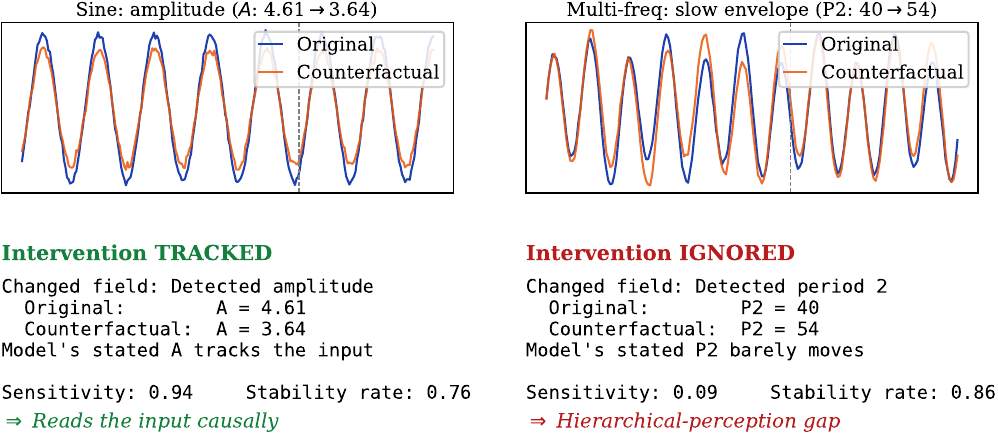}}
\caption{Counterfactual probe, success versus failure. On the left,
changing the Sine amplitude shifts the model's stated detected amplitude
to match ($\mathrm{Sensitivity}{=}0.94$), a causal reading of the input. On the
right, swapping Multi-freq's slow envelope (period 2) leaves the
stated value almost unchanged ($\mathrm{Sensitivity}{=}0.09$), localizing the
hierarchical-perception gap.}
\label{fig:cf}

\end{figure*}

\textbf{Metric configuration.}
\label{sec:appendix_metricconfig}
Table~\ref{tab:tol} lists the per-field tolerances of the $\mathrm{match}$ indicator,
which Fidelity, Sensitivity, and Stability all share. A zero pair requires an exact
match, and the ranges reflect per-pattern differences in a field's natural scale.
Table~\ref{tab:cf_spec} lists the counterfactual interventions of the Sensitivity
probe. Each new value is drawn from the parameter's generative range and redrawn
until it differs from the original by a minimum margin, which keeps every
counterfactual a real change, and the intervened field is scored with the same
tolerances as Fidelity.

\begin{table}[h]
\centering
\small
\setlength{\tabcolsep}{5pt}
\renewcommand{\arraystretch}{1.15}
\begin{tabular}{@{}lcc@{}}
\toprule
\textbf{Field} & $a_f$ & $r_f$ \\
\midrule
Period or index      & $0$            & $0$ \\
Event time           & $0$            & $0$ \\
Categorical label    & $0$            & $0$ \\
Changepoint index    & $2$            & $0$ \\
Amplitude            & $0.10$--$0.15$ & $0.10$--$0.15$ \\
Phase                & $0.3$          & $0$ \\
Slope                & $0.01$         & $0.2$ \\
Intercept            & $0.5$          & $0.2$ \\
AR coefficient       & $0.05$         & $0$ \\
Noise std            & $0.05$         & $0.3$ \\
Last value           & $0.10$--$0.20$ & $0.1$ \\
\bottomrule
\end{tabular}
\caption{Per-field tolerances for the $\mathrm{match}$ indicator, with $a_f$
absolute and $r_f$ relative.}
\label{tab:tol}
\end{table}

\begin{figure*}[!t]
\centering
\setlength{\tabcolsep}{2pt}\renewcommand{\arraystretch}{1.15}\small
\begin{tabular}{@{}l | >{\raggedright\arraybackslash}m{0.46\textwidth} | >{\raggedright\arraybackslash}m{0.27\textwidth} | c@{}}
\toprule
\textbf{Configuration} & \textbf{What it is} & \textbf{Trained on} & \textbf{\# Models} \\
\midrule
Explicit routing & One model. The first emitted token is the pattern class, then a class-conditional reasoning chain and forecast & All patterns $+$ Unknown, with a pattern-class prefix & 1 \\
\midrule
Implicit routing & The same single model without the class token, so routing is learned implicitly & All patterns $+$ Unknown & 1 \\
\midrule
Single-pattern & A separate specialist per pattern, so an external caller must pick the model & One pattern & 5 \\
\bottomrule
\end{tabular}
\captionof{table}{Our default model and its two ablations. Routing is how the model
decides which output schema to emit. Explicit routing (our default) makes routing
an explicit first-token step and is the only configuration deployable on an
arbitrary input. Implicit routing learns routing implicitly. Single-pattern
trains a separate specialist per pattern with no cross-pattern sharing and needs
an external caller, isolating the value of joint training rather than serving as
a deployable system.}
\label{tab:variants}

\adjustbox{max width=\textwidth}{\includegraphics{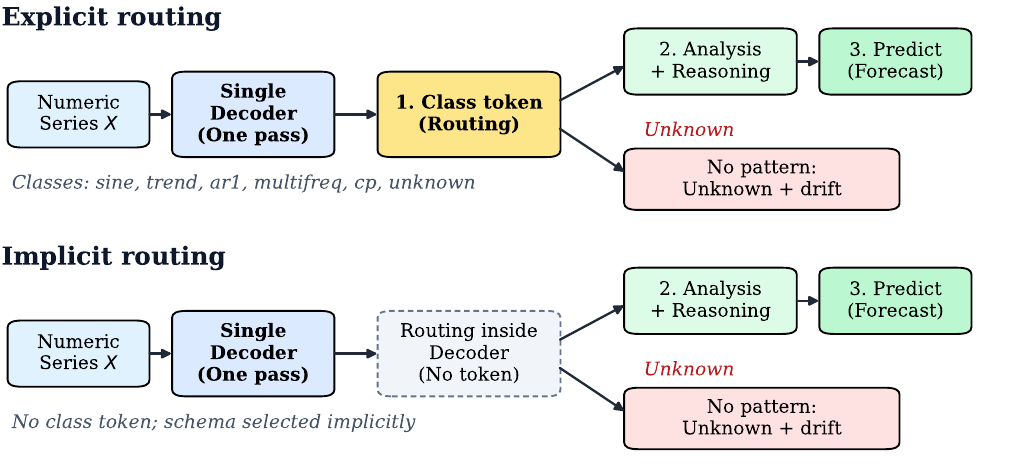}}
\captionof{figure}{The two ReasonCast routing pipelines. A single decoder produces the
class-conditional reasoning chain and the numeric prediction in one autoregressive pass,
with a no-pattern branch for inputs matching no primitive. [Top] Explicit routing emits
the pattern class as the first token. [Bottom] Implicit routing selects the schema inside
the decoder.}
\label{fig:flow}
\end{figure*}

\begin{table}[h]
\centering
\small
\setlength{\tabcolsep}{5pt}
\renewcommand{\arraystretch}{1.15}
\begin{tabular}{@{}l|l@{}}
\toprule
\textbf{Intervened} & \textbf{Counterfactual value} \\
\midrule
\multicolumn{2}{@{}l}{\textit{Sine}} \\
Period    & $\mathrm{int}[4,50]$, $\neq$ orig \\
Amplitude & $U(0.5,5.0)$, $|\Delta|\!\ge\!0.5$ \\
Phase     & $U(0,2\pi)$, $|\Delta|\!\ge\!0.5$ \\
\midrule
\multicolumn{2}{@{}l}{\textit{Trend}} \\
Period    & $\mathrm{int}[4,50]$, $\neq$ orig \\
Slope     & $U(-0.05,0.05)$, $|\Delta|\!\ge\!0.02$ \\
Intercept & $U(-2,2)$, $|\Delta|\!\ge\!1.0$ \\
\midrule
\multicolumn{2}{@{}l}{\textit{AR(1)}} \\
Alpha & $U(-0.95,0.95)$, $|\Delta|\!\ge\!0.3$ \\
\midrule
\multicolumn{2}{@{}l}{\textit{Multi-freq}} \\
Period$_1$    & $\mathrm{int}[3,20]$, $\neq$, $2p_1\!<\!p_2$ \\
Period$_2$    & $\mathrm{int}[2p_1,60]$, $\ge\!2p_1$ \\
Amplitude$_1$ & $U(0.5,3.0)$, $|\Delta|\!\ge\!0.5$ \\
\midrule
\multicolumn{2}{@{}l}{\textit{Changepoint}} \\
Changepoint $t$ & $\mathrm{int}[0.25N,0.85N]$, $|\Delta|\!\ge\!0.1N$ \\
Segment 2 slope & $U(-0.1,0.1)$, $|\Delta|\!\ge\!0.03$ \\
\bottomrule
\end{tabular}
\caption{Counterfactual interventions. $N$ is the context length and $U$ denotes a
uniform draw over the stated range.}
\label{tab:cf_spec}
\end{table}

\begin{table}[t]
\centering
\small
\setlength{\tabcolsep}{3pt}
\renewcommand{\arraystretch}{1.1}

\begin{tabular}{@{}lccl@{}}
\toprule
\textbf{Reasoning field} & $\hat{v}$ & $\hat{v}'$ & \textbf{Contributes} \\
\midrule
Detected period$^{\dagger}$ & $36$ & $40$ & Sensitivity: $\mathrm{match}(40,40)=1$ \\
Detected amplitude & $3.65$ & $3.66$ & Stability: $\mathrm{match}(3.66,3.65)=1$ \\
Detected phase & $2.31$ & $2.33$ & Stability: $\mathrm{match}(2.33,2.31)=1$ \\
\midrule
\multicolumn{4}{@{}l}{Sensitivity $=1$, \quad Stability $=(1+1)/2=1$} \\
\bottomrule
\end{tabular}
\caption{A counterfactual probe on one Sine pair. $^{\dagger}$ is the intervened
parameter (period $36\!\to\!40$), and $\hat{v}$, $\hat{v}'$ are the model's fields on both
inputs.}
\label{tab:cf_worked}
\end{table}

\section{Counterfactual Probe: A Worked Example}
\label{sec:appendix_cfprobe}

Table~\ref{tab:cf_worked} runs the probe on a Sine sample where only the period is
changed, $\theta_j\!:36\!\to\!40$, holding everything else fixed. The model reads
each reasoning field twice, once on the base input ($\hat{v}$) and once on the
counterfactual ($\hat{v}'$). The intervened field, the detected period, follows the
change to $40$, so $\mathrm{Sensitivity}=\mathrm{match}(40,40)=1$, and the two untouched
fields stay within tolerance across the pair, so $\mathrm{Stability}=2/2=1$. The model
therefore tracked the intervened parameter while leaving the two untouched fields
in place. Figure~\ref{fig:cf} contrasts such a tracked intervention with an ignored
one, where the model leaves a Multi-freq slow envelope almost unchanged.

\section{ReasonCast Pipeline}
\label{sec:appendix_pipeline}

Figure~\ref{fig:flow} shows the two single-model routing pipelines end to end, and
Table~\ref{tab:variants} compares the three configurations we study in this work.
A single decoder produces the entire output in one autoregressive pass. It first emits a
class-conditional reasoning chain, split into an INPUT ANALYSIS block that states the
detected structure of the series and a REASONING block that derives the forecast from
that structure, and then emits the numeric PREDICTION block. When an input matches none
of the five primitives, a no-pattern branch fires instead, and the model reports low
pattern-match confidence and extrapolates the recent local trend rather than forcing a
primitive schema onto the input.

The two variants differ only in how the pattern class is routed. Explicit routing emits
the class as the first generated token, so routing is a visible step that conditions all
that follows. Implicit routing emits no class token and selects the schema inside the
decoder. Both share the factorization
$p_\theta(o_{\mathrm{rsn}}\mid X)\,p_\theta(o_{\mathrm{gen}}\mid o_{\mathrm{rsn}}, X)$, in
which the reasoning is produced first and the generation is conditioned on it. At 3B the
two variants reach the same error and nearly identical Fidelity and Consistency, as
Table~\ref{tab:qwen_full} reports. The class token makes the routing visible without
changing accuracy.

\begin{table*}[!t]
\centering
\setlength{\tabcolsep}{6pt}
\renewcommand{\arraystretch}{1.15}
\small
\begin{tabular}{@{}l c | c c c c@{}}
\toprule
\textbf{Variant} & \textbf{Scale} & \textbf{Error}~$\downarrow$ & \textbf{Fidelity}~$\uparrow$ & \textbf{Consistency}~$\uparrow$ & \textbf{Sensitivity}~$\uparrow$ \\
\midrule
\multirow{3}{*}{Explicit routing} & 0.5B & 0.379 & 0.804 & 0.449 & 0.662 \\
 & 1.5B & 0.259 & 0.860 & 0.528 & 0.718 \\
 & 3B & 0.233 & 0.900 & 0.620 & 0.794 \\
\midrule
\multirow{3}{*}{Implicit routing} & 0.5B & 0.339 & 0.778 & 0.379 & -- \\
 & 1.5B & 0.259 & 0.863 & 0.542 & -- \\
 & 3B & 0.233 & 0.894 & 0.622 & -- \\

\bottomrule
\end{tabular}
\caption{Full performance grid for the Qwen backbone, with every metric for Qwen2.5-Instruct across scale and the two single-model variants (Explicit and Implicit routing), all under full fine-tuning. Error is the mean forecast MAE. This table reports 200 test samples per pattern, versus the full 1{,}000 used by the main-text Tables~\ref{tab:forecast} and~\ref{tab:reasoning_compare}.}
\label{tab:qwen_full}
\end{table*}

\section{Baseline Details}
\label{sec:appendix_baselines_sec}

\subsection{Baseline Catalog and Configuration}
\label{sec:appendix_baseline_catalog}

We compare against a wide set of baselines spanning the four axes of
Table~\ref{tab:axes}. Table~\ref{tab:baselines} lists them by category with the
capabilities each offers. Not all produce reasoning or no-pattern detection, but
all produce a forecast, so the forecast metric is comparable throughout.

\begin{table}[h]
\centering
\setlength{\tabcolsep}{3pt}
\renewcommand{\arraystretch}{1.12}

\small
\begin{tabular}{@{}lccc@{}}
\toprule
\textbf{Method} & \textbf{Forecast} & \textbf{Reasoning} & \textbf{No-pattern} \\
\midrule
\multicolumn{4}{@{}l}{\textit{Trivial}} \\
Naive last            & \checkmark & --      & -- \\
Seasonal naive       & \checkmark & --      & -- \\
Linear extrapolation & \checkmark & --      & -- \\
\midrule
\multicolumn{4}{@{}l}{\textit{TS forecasting models}} \\
DLinear      & \checkmark & -- & -- \\
PatchTST     & \checkmark & -- & -- \\
iTransformer & \checkmark & -- & -- \\
TimeXer      & \checkmark & -- & -- \\
\midrule
\multicolumn{4}{@{}l}{\textit{LLM (4 backbones)}} \\
Few-shot & \checkmark & Partial    & -- \\
Fine-tuned         & \checkmark & \checkmark & -- \\
\midrule
\multicolumn{4}{@{}l}{\textit{LLM $+$ ReasonCast}} \\
Numeric-only SFT & \checkmark & --         & -- \\
Single-pattern   & \checkmark & \checkmark & -- \\
Implicit routing & \checkmark & \checkmark & \checkmark \\
Explicit routing & \checkmark & \checkmark & \checkmark \\
\bottomrule
\end{tabular}

\caption{Comparison of baselines. Only our methods cover all three capabilities
simultaneously.}
\label{tab:baselines}
\end{table}

\textbf{Trivial baselines.} Naive-last repeats $x(N{-}1)$ for all future steps.
Seasonal-naive estimates a period from the autocorrelation function and repeats it, and
an oracle variant instead uses the ground-truth period as an upper bound. Linear
extrapolation fits a trend to the last 24 points and extends that trend over the
horizon.

\textbf{TS forecasting models.} DLinear \citep{Zeng2023DLinear}, PatchTST
\citep{Nie2023PatchTST}, iTransformer \citep{Liu2024iTransformer}, and TimeXer
\citep{Wang2024TimeXer}, all trained per-pattern on our 12k training set with
matched optimizer and 20 epochs, each with a 200-step input window and 100-step
output head.

\textbf{LLM baselines.} We benchmark four instruction-tuned LLM backbones
spanning the Qwen, Llama, Phi, and Gemma families
(Qwen2.5-7B-Instruct~\citep{Yang2024Qwen25},
Llama-3.1-8B-Instruct~\citep{Grattafiori2024Llama3},
Phi-3.5-mini-instruct~\citep{Abdin2024Phi3}, and
Gemma-2-9B-it~\citep{Riviere2024Gemma2}). Each is evaluated both few-shot, with
$k{=}2$ in-context examples sampled from the training set and no weight updates,
and fine-tuned on our reasoning chains with the same recipe as ReasonCast
(low-rank adaptation~\citep{Hu2022LoRA} for the 7B--9B backbones).

\textbf{Our ablations.} Numeric SFT uses the same base model as Single-pattern but
trained to emit only the PREDICTION block, with no reasoning chain. This
isolates the cost or benefit of attaching reasoning to forecasting.

\subsection{Evaluation Protocol}
\label{sec:appendix_eval_protocol}

Unless noted, all forecast metrics use $n{=}200$ test samples per pattern. The
reasoning metrics (Fidelity, Consistency, and Sensitivity) are reported only for
methods that emit a reasoning chain, since every other baseline scores $0$ on them
by construction, so we omit those zero columns. ReasonCast is trained for one epoch
per pattern or one epoch on the 72k union (no-pattern
and Explicit routing) at 0.5B, 1.5B, and 3B, with 3B as the default. DLinear and
PatchTST are trained per pattern for 20 epochs with a matched optimizer.

\section{Explicit vs.\ Implicit Routing}
\label{sec:appendix_qwen_full}

Table~\ref{tab:qwen_full} records every metric for the Qwen backbone across the
three scales (0.5B, 1.5B, 3B) and the two single-model routing variants, Explicit
and Implicit routing, all under full fine-tuning. Sensitivity is reported only for
Explicit routing, since it needs the paired counterfactual runs of
Appendix~\ref{sec:appendix_cfprobe} and we ran those for the default variant only, so
the Implicit rows carry a dash. Two patterns hold across the grid.
First, every metric improves with scale: forecast error falls while Fidelity and
Consistency rise from 0.5B to 3B. Second, the two
routing variants track each other closely at every scale, and at 3B they reach the
same forecast error and nearly identical Fidelity and Consistency. This confirms the
main-text result that the gain comes from the reasoning itself rather than from how the
pattern class is exposed, and emitting the class as an explicit token or selecting it
implicitly inside the decoder are interchangeable in practice.

\section{Grounding Ablations}
\label{sec:appendix_grounding}

\begin{figure}[h]
\centering
\adjustbox{max width=\columnwidth}{\includegraphics{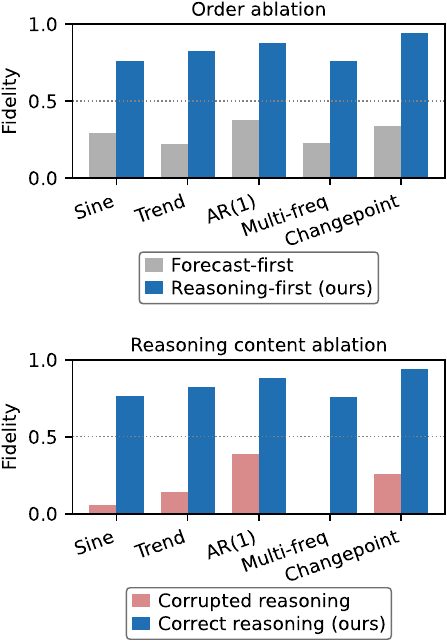}}
\caption{Two ways the reasoning loses its grounding. [Top] The order ablation emits
the forecast before the reasoning. [Bottom] The reasoning content ablation trains on
the reasoning of another sample. Both collapse Fidelity relative to the grounded
model.}
\label{fig:grounding}
\end{figure}

This section visualizes the two ablations that break the reasoning's grounding, each
also reported numerically in the main text. The first is the order ablation reported in
Table~\ref{tab:order}. Emitting the forecast before the reasoning turns the explanation
into a post-hoc rationalization and collapses Fidelity on every pattern. The second is
the reasoning content ablation reported in Table~\ref{tab:corrupt}. Replacing each
sample's reasoning target with the reasoning of another sample keeps the required
format but leaves the fields describing a different input. This collapses Fidelity and
Consistency on every pattern, and it degrades the forecast as well, even though the
forecast targets were left untouched. Figure~\ref{fig:grounding} plots both
effects, and the corrupted Multi-freq model degenerates far enough to emit nothing
parseable. Emitting the reasoning first and supervising it with the correct content are
therefore both necessary for the reasoning to match what the model predicts.

\section{Symbolic Regression Baseline}
\label{sec:appendix_symbolic}

\textbf{Data.} We evaluate the symbolic-regression baseline~\citep{SchmidtLipson2009,SINDy2016} on the ReasonTS-Bench
forecasting test sets, using the same five test splits and the same forecast MAE as
Table~\ref{tab:forecast}. The naive-last predictor reproduces its published MAE on these
splits, so the two are on an identical scale. We report the full 1{,}000 test samples per
pattern.

\textbf{Method.} We give the baseline the strongest possible footing: its candidate
library is the benchmark's own five generative forms, so it only fits parameters and
selects among the right hypotheses rather than searching an open expression space. This
is a library-restricted, SINDy-style fit~\citep{SINDy2016}. For each context window we
fit every form by least squares, select the form with the lowest BIC, and extrapolate it
over the horizon, and the recovered parameters are the structural fields credited in
Table~\ref{tab:symbolic}. It emits no natural-language reasoning, so by construction it
offers none of the language-side capabilities that Table~\ref{tab:symbolic} credits to
ReasonCast.

\section{Multi-Seed Standard Deviation}
\label{sec:appendix_forecast_std}

Table~\ref{tab:forecast_std} reports the default model's forecast error as mean and
standard deviation over 3 random seeds, which differ only in the random initialization
and the training-data order. The standard deviations stay far below the gap between
ReasonCast and its baselines.

\begin{table}[h]
\centering
\setlength{\tabcolsep}{4pt}
\renewcommand{\arraystretch}{1.15}
\small
\begin{tabular}{@{}l|cc@{}}
\toprule
\textbf{Pattern} & \textbf{MAE} & \textbf{MSE} \\
\midrule
Sine  & $0.208 \pm 0.011$ & $0.139 \pm 0.013$ \\
Trend & $0.360 \pm 0.016$ & $0.343 \pm 0.022$ \\
AR    & $0.181 \pm 0.005$ & $0.073 \pm 0.004$ \\
MF    & $0.256 \pm 0.014$ & $0.225 \pm 0.018$ \\
CP    & $0.177 \pm 0.009$ & $0.145 \pm 0.011$ \\
\midrule
Avg.  & $0.236 \pm 0.008$ & $0.185 \pm 0.010$ \\
\bottomrule
\end{tabular}
\caption{Forecast error of the default 3B model (Qwen2.5-3B with ReasonCast) across
the five primitives, as mean $\pm$ standard deviation over 3 random seeds.}
\label{tab:forecast_std}
\end{table}

\section{Capacity Scaling}
\label{sec:appendix_scaling}

To measure how backbone size affects ReasonCast, we train it at three scales and
report all four metrics in Table~\ref{tab:reasoning}. Every metric improves with
scale, and the gain is largest on the patterns with the most structure to recover,
whereas AR is already near its ceiling at 0.5B. The results indicate that reasoning
quality grows with capacity, and we adopt the 3B model as our default.

\begin{table*}[t]
\centering
\small
\setlength{\tabcolsep}{4.5pt}
\begin{tabular}{l|ccc|ccc|ccc|ccc}
\toprule
& \multicolumn{3}{c|}{\textbf{Error}~$\downarrow$} & \multicolumn{3}{c|}{\textbf{Fidelity}~$\uparrow$} & \multicolumn{3}{c|}{\textbf{Consistency}~$\uparrow$} & \multicolumn{3}{c}{\textbf{Sensitivity}~$\uparrow$} \\
\cmidrule(lr){2-4}\cmidrule(lr){5-7}\cmidrule(lr){8-10}\cmidrule(lr){11-13}
\textbf{Pattern} & \textbf{0.5B} & \textbf{1.5B} & \textbf{3B} & \textbf{0.5B} & \textbf{1.5B} & \textbf{3B} & \textbf{0.5B} & \textbf{1.5B} & \textbf{3B} & \textbf{0.5B} & \textbf{1.5B} & \textbf{3B} \\
\midrule
Sine  & 0.221 & \textbf{0.203} & \underline{0.208} & 0.821 & \underline{0.905} & \textbf{0.926} & 0.596 & \underline{0.703} & \textbf{0.753} & 0.870 & \underline{0.940} & \textbf{0.960} \\
Trend & 0.484 & \underline{0.393} & \textbf{0.360} & 0.795 & \underline{0.872} & \textbf{0.898} & 0.206 & 0.330 & \textbf{0.414} & 0.770 & \underline{0.850} & \textbf{0.880} \\
AR    & 0.185 & \underline{0.183} & \textbf{0.181} & 0.859 & \textbf{0.868} & \underline{0.864} & 0.997 & \textbf{0.998} & \textbf{0.998} & 0.480 & \underline{0.520} & \textbf{0.530} \\
MF    & 0.439 & \underline{0.345} & \textbf{0.256} & 0.733 & \underline{0.802} & \textbf{0.883} & 0.181 & \underline{0.247} & \textbf{0.445} & 0.700 & \underline{0.690} & \textbf{0.820} \\
CP    & 0.227 & \underline{0.188} & \textbf{0.177} & 0.809 & \underline{0.856} & \textbf{0.924} & 0.304 & \underline{0.356} & \textbf{0.458} & 0.490 & \underline{0.590} & \textbf{0.780} \\
\midrule
Average & 0.311 & \underline{0.262} & \textbf{0.236} & 0.803 & \underline{0.861} & \textbf{0.899} & 0.457 & 0.527 & \textbf{0.613} & 0.662 & \underline{0.718} & \textbf{0.794} \\
\bottomrule
\end{tabular}
\caption{Reasoning correctness across scales. All four metrics improve as ReasonCast grows from 0.5B to 3B.}
\label{tab:reasoning}
\end{table*}

\section{Ground-Truth Reasoning Chains}
\label{sec:appendix_examples}

For each pattern we show one real sample's complete ground-truth reasoning
chain, exactly as emitted by the generator (Section~\ref{sec:bench}), with no
simplification. Each chain has the three labeled blocks INPUT ANALYSIS,
REASONING, and PREDICTION, and every field is computed in closed form from the
generative parameters, so it is verifiable. We show the smallest-horizon sample
of each pattern for compactness, with its series plotted above the chain (blue is
the input context, red is the target).

\begin{figure}[h]
\centering
\adjustbox{max width=\columnwidth}{\includegraphics{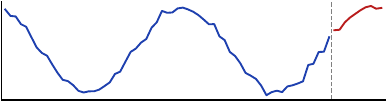}}\par\smallskip
\begin{Verbatim}[fontsize=\small,frame=single,framesep=3pt]
INPUT ANALYSIS:
  observed_length: 63
  detected_period: 36
  detected_amplitude: 3.71
  detected_phase: 1.9431
  last_value: x(62) = 1.1407
  current_phase_in_cycle: 0.031
  current_state: near_trough

REASONING:
  pattern_type: pure_sinusoidal
  rule: x(t) = 3.71 * sin(2pi*t/36
              + 1.9431)
  next_peak_at: t = 70
  next_trough_at: t = 88
  next_cycle_completes_at: t = 98

PREDICTION:
  t=63: 1.35
  t=64: 1.929
  t=65: 2.45
  t=66: 2.897
  t=67: 3.255
  t=68: 3.515
  t=69: 3.668
  t=70: 3.709
  t=71: 3.638
  t=72: 3.456
\end{Verbatim}
\caption{Sine.}
\end{figure}

\begin{figure}[t]
\centering
\adjustbox{max width=\columnwidth}{\includegraphics{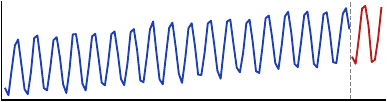}}\par\smallskip
\begin{Verbatim}[fontsize=\small,frame=single,framesep=3pt]
INPUT ANALYSIS:
  observed_length: 108
  detected_period: 6
  detected_amplitude: 4.47
  detected_phase: 4.0242
  detected_slope: 0.0452
  detected_intercept: 1.013
  last_value: x(107) = 6.8411
  current_phase_in_cycle: 0.474
  current_state: near_peak

REASONING:
  pattern_type: sine_plus_linear_trend
  rule: x(t) = 4.47 * sin(2pi*t/6 + 4.0242)
              + 0.0452*t + 1.013
  trend_direction: rising
  next_peak_at: t = 112
  next_trough_at: t = 109

PREDICTION:
  t=108: 2.442
  t=109: 1.755
  t=110: 5.253
  t=111: 9.483
  t=112: 10.26
  t=113: 6.853
  t=114: 2.713
  t=115: 2.026
  t=116: 5.524
  t=117: 9.754
\end{Verbatim}
\caption{Sine $+$ Trend.}
\end{figure}

\begin{figure}[t]
\centering
\adjustbox{max width=\columnwidth}{\includegraphics{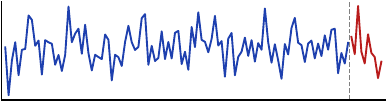}}\par\smallskip
\begin{Verbatim}[fontsize=\small,frame=single,framesep=3pt]
INPUT ANALYSIS:
  observed_length: 104
  detected_alpha: 0.045
  detected_noise_std: 0.088
  last_value: x(103) = 0.0215
  regime: monotonic_decay

REASONING:
  pattern_type: ar1
  rule: x(t) = 0.045 * x(t-1) + eps
  long_run_mean: 0
  half_life_steps: 0.22
  forecast_rule: x_hat(N+k)
                 = alpha^(k+1) * x(N-1)
                 = 0.045^(k+1) * 0.0215

PREDICTION:
  t=104: 0.001
  t=105: 0.0
  t=106: 0.0
  t=107: 0.0
  t=108: 0.0
  t=109: 0.0
  t=110: 0.0
  t=111: 0.0
  t=112: 0.0
  t=113: 0.0
\end{Verbatim}
\caption{AR(1).}
\end{figure}

\begin{figure}[t]
\centering
\adjustbox{max width=\columnwidth}{\includegraphics{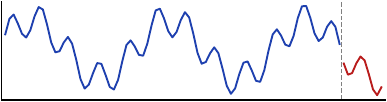}}\par\smallskip
\begin{Verbatim}[fontsize=\small,frame=single,framesep=3pt]
INPUT ANALYSIS:
  observed_length: 81
  detected_period_1: 7
  detected_period_2: 33
  detected_amplitude_1: 1.05
  detected_amplitude_2: 1.84
  detected_phase_1: 0.2691
  detected_phase_2: 0.3848
  last_value: x(80) = 0.3595

REASONING:
  pattern_type: two_frequency_superposition
  rule: x(t) = 1.05*sin(2pi*t/7 + 0.2691)
              + 1.84*sin(2pi*t/33 + 0.3848)
  dominant_component: component_2
  amplitude_ratio_A1_over_A2: 0.571

PREDICTION:
  t=81: -0.873
  t=82: -1.574
  t=83: -1.467
  t=84: -0.864
  t=85: -0.43
  t=86: -0.672
  t=87: -1.554
  t=88: -2.513
  t=89: -2.887
  t=90: -2.404
\end{Verbatim}
\caption{Multi-frequency.}
\end{figure}

\begin{figure}[t]
\centering
\adjustbox{max width=\columnwidth}{\includegraphics{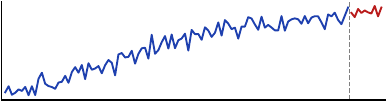}}\par\smallskip
\begin{Verbatim}[fontsize=\small,frame=single,framesep=3pt]
INPUT ANALYSIS:
  observed_length: 104
  detected_changepoint_t: 64
  segment_1_slope: 0.0504
  segment_1_intercept: 2.207
  segment_2_slope: 0.0176
  segment_2_intercept: 4.3062
  last_value: x(103) = 6.5349
  current_segment: segment_2

REASONING:
  pattern_type:
      piecewise_linear_with_changepoint
  rule_seg1: x(t) = 0.0504*t + 2.207,
             valid for t <= 64
  rule_seg2: x(t) = 0.0176*t + 4.3062,
             valid for t > 64
  regime_after_changepoint: decelerating
  forecast_uses_segment_2: True
      (since N-1=103 > t*=64)

PREDICTION:
  t=104: 6.137
  t=105: 6.154
  t=106: 6.172
  t=107: 6.189
  t=108: 6.207
  t=109: 6.225
  t=110: 6.242
  t=111: 6.26
  t=112: 6.277
  t=113: 6.295
\end{Verbatim}
\caption{Changepoint.}
\end{figure}

\end{document}